\documentclass{article}

\PassOptionsToPackage{numbers, sort&compress}{natbib}

\usepackage[main, final]{neurips_2026}

\usepackage[utf8]{inputenc} 
\usepackage[T1]{fontenc}    
\usepackage{hyperref}       
\usepackage{url}            
\usepackage{booktabs}       
\usepackage{amsfonts}       
\usepackage{nicefrac}       
\usepackage{microtype}      
\usepackage{xcolor}         
\usepackage{pifont}
\newcommand{\cmark}{\textcolor{green}{\ding{51}}}
\newcommand{\xmark}{\textcolor{red}{\ding{55}}}

\usepackage{amsmath,amssymb,bm}
\usepackage{booktabs}
\usepackage{tikz}
\usetikzlibrary{arrows.meta,positioning,calc,fit}

\usepackage{booktabs} \usepackage{multirow} 
\usepackage{makecell}
\definecolor{MyRed}{RGB}{192,0,0}
\definecolor{MyBlue}{RGB}{0,92,175}

\newcommand{\best}[1]{\textbf{\textcolor{MyRed}{#1}}}
\newcommand{\second}[1]{\textbf{\textcolor{MyBlue}{#1}}}

\usepackage[table]{xcolor}
\newcommand{\cao}[1]{\textcolor{black}{#1}}

\usepackage{amsmath,amssymb,bm}
\usepackage{algorithm}
\usepackage{algpseudocode}
\usepackage{wrapfig}
\usepackage{enumitem}
\title{\cao{Efficient One-Step Diffusion Restoration Model with Compact Token Compression and Linear Attention}}

\author{%
  \!\!\!\!\!\!Bingtian Qiao$^{1,3}$\thanks{Work done during an internship at Shanghai Jiao Tong University.}\quad 
  Yue Shi$^{2,\dagger}$\quad 
  Yingjie Zhou$^1$\quad 
  Yong Guo$^1$\quad 
  Guangtao Zhai$^{1,2}$\quad 
  Jiezhang Cao$^{1}$\thanks{Corresponding author. Email: \texttt{caojiezhang@sjtu.edu.cn,shiyue@pjlab.org.cn}} \\
  $^1$Shanghai Jiao Tong University~~~ 
  $^2$Shanghai AI Laboratory~~~ 
  $^3$Fuzhou University 
  \\
}

\begin{document}

\maketitle

\begin{figure}[htb]
    \vspace{-3mm}
	\centering
	\includegraphics[width=\linewidth]{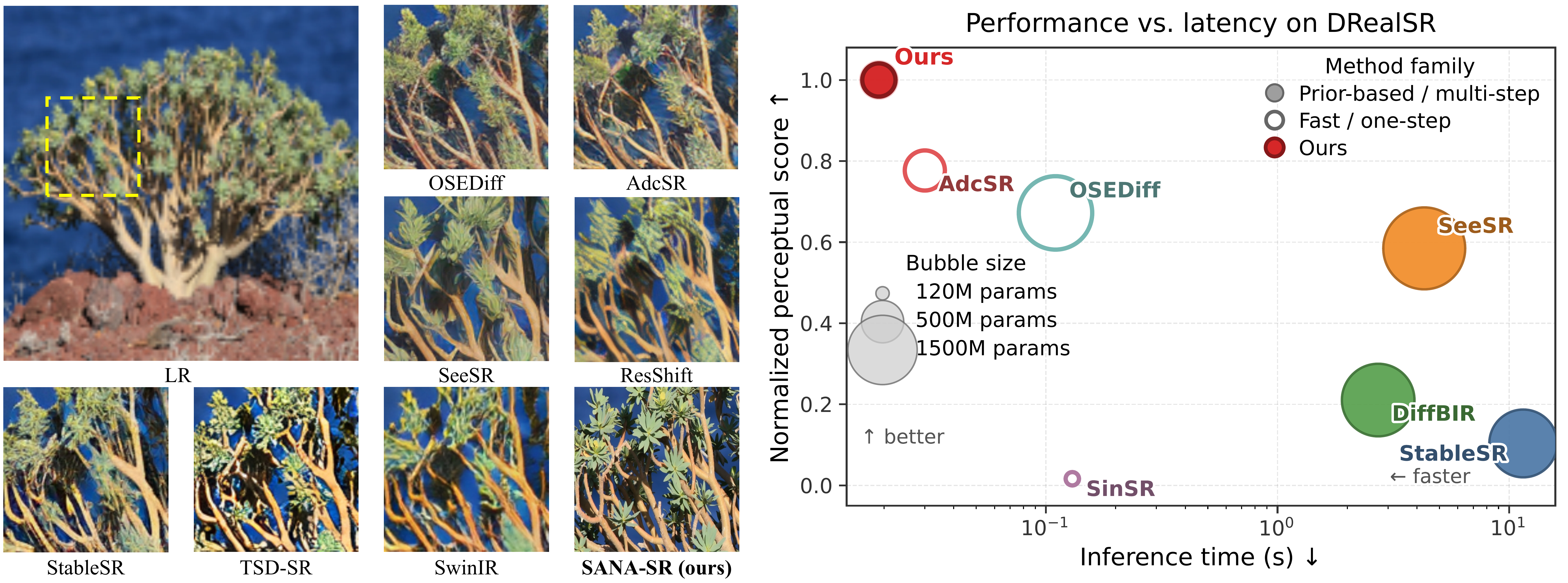}
    \vspace{-7mm}
	\caption{\textbf{SANA-SR achieves a strong quality--efficiency trade-off for real-world image super-resolution. Left:} qualitative comparison on a real LR input against seven baselines, the yellow box is zoom region. \textbf{Right:} DRealSR scatter of normalized perceptual score and inference time; marker color encodes method family and size scales with parameters. SANA-SR yields a best perceptual at the lowest latency.}
	\label{fig:framework}
\end{figure}

\begin{abstract}
Real-world image super-resolution aims to recover high-quality images from complex and unknown real-world degradations. 
\cao{However, existing generative Real-ISR methods largely inherit the dense latent representations and quadratic-cost global modeling paradigm developed for high-resolution image synthesis, causing computation, memory usage, and inference latency to scale unfavorably with resolution and thus limiting practical deployment. We argue that the key bottleneck lies not in insufficient restoration priors, but in excessive token redundancy and costly token interactions during high-resolution restoration. Motivated by this observation, we revisit Real-ISR from the perspectives of compact latent representation and linear-complexity modeling, and propose SANA-SR, an efficient one-step restoration framework. Specifically, SANA-SR employs a deep compression autoencoder with a \textbf{$\bf{32{\times}}$ compression ratio} to drastically reduce latent tokens while preserving restoration-relevant structures and textures. On top of this compact latent space, we introduce a linear-attention DiT with LoRA fine-tuning, enabling efficient high-resolution restoration with linear-complexity token mixing. Extensive experiments on all benchmark datasets demonstrate that SANA-SR achieves highly competitive and often superior quantitative performance against existing methods, while restoring clearer and more realistic textures. Moreover, after pruning, the deployed model runs in \textbf{0.019s} with \textbf{407.95G MACs} and \textbf{344M parameters}, highlighting its strong potential for practical mobile deployment.}
\end{abstract}

\section{Introduction}
\label{sec:intro}
\vspace{-2mm}
Real-world image super-resolution (Real-ISR) \citep{wang2020deep, cai2019toward} aims to recover high-quality images from low-quality observations degraded by unknown blur, noise, compression, and sensor artifacts \citep{wang2024stablesr}. 
\cao{Despite recent progress, existing methods still face a fundamental trade-off between restoration quality and efficiency \citep{lugmayr2020ntire,wang2020deep,ignatov2025quantized,xiao2026reversible}: approaches that recover more realistic texture and rely on heavy generative models and costly high-resolution (HR) processing, while efficient models tend to compromise fine-detail reconstruction and visual realism. Bridging the gap between restoration quality and efficiency therefore remains a central challenge for practical super-resolution.}

\cao{Recent years have witnessed the growing success of generative restoration methods for Real-ISR~\citep{wang2024stablesr, lin2024diffbir, wu2024seesr, wu2024osediff, chen2025adcsr, dong2025tsdsr}. Compared with purely regression-based approaches~\citep{dong2014learning, lim2017enhanced, zhang2018image, liang2021swinir, chen2023activating}, generative models better address the severe ill-posedness of real-world degradations by synthesizing more plausible textures and visually realistic details. However, as shown in Table\ref{tab:paradigm_comparison}, existing methods still occupy only part of the design space. SD-based \citep{rombach2022high} and FLUX-based \citep{flux2024,labs2025flux1kontextflowmatching} one-step methods avoid iterative sampling, yet they retain dense 8$\times$ latent representations with $\sim$4K tokens and quadratic-cost attention. In contrast, linear-attention methods \citep{katharopoulos2020transformers,xie2025sana} reduce token interaction complexity and benefit from stronger latent compression, but representative models such as LinearSR \citep{li2026linearsr} are trained from scratch and remain multi-step. As a result, none of these families simultaneously achieves one-step inference, compact 32$\times$ latent compression, linear-complexity attention, and lightweight deployment. This gap suggests that the main inefficiency of current Real-ISR systems lies not only in sampling depth, but more fundamentally in the dense high-resolution modeling paradigm inherited from image synthesis.}

\cao{This observation motivates us to revisit Real-ISR from two important perspectives: compact latent representation and efficient token interaction. To this end, we propose SANA-SR, an efficient one-step restoration framework, as shown as Fig.~\ref{fig:framework}. Specifically, we first leverage a degradation-aware deep compression autoencoder with a $32{\times}$ compression ratio, which significantly reduces the number of latent tokens while preserving restoration-relevant structural and textural information. Building upon this compact latent space, we introduce a DiT restoration backbone \citep{peebles2023scalable} with linear attention \citep{katharopoulos2020transformers, xie2025sana}, which replaces quadratic-cost token interactions with linear-complexity modeling. To further adapt the model efficiently, we adopt LoRA fine-tuning \citep{hu2022lora}, enabling effective restoration learning with moderate training overhead. 
Extensive experiments on standard Real-ISR benchmarks demonstrate that SANA-SR achieves highly competitive and often superior quantitative performance compared with existing methods, while restoring clearer and more realistic textures. More importantly, our method exhibits strong deployment efficiency: after pruning, the deployed model runs in 0.019 s with 407.95G MACs and 344M parameters, showing strong potential for practical mobile applications.}

Our main contributions are summarized as follows:
\begin{itemize}[nosep,leftmargin=0.3cm]
    \setlength{\itemsep}{0pt}
    \setlength{\parsep}{0pt}
    \setlength{\parskip}{4pt}
    \item \cao{We identify token redundancy and costly high-resolution token interactions as a key yet underexplored bottleneck in generative Real-ISR, and revisit the task from the perspectives of compact latent representation and linear-complexity modeling.}
    \item \cao{We propose SANA-SR, an efficient one-step Real-ISR framework that includes a $32{\times}$ degradation-aware deep compression autoencoder, linear-attention DiT for linear complexity, and LoRA fine-tuning to enable efficient high-resolution restoration.}
    \item \cao{Our SANA-SR achieves a favorable efficiency-quality trade-off, delivering highly competitive and often superior quantitative results with clearer and more realistic textures across benchmark datasets, while also showing strong potential for practical mobile deployment.}
\end{itemize}


\section{Related Work}
\label{sec:related}
\vspace{-2mm}

\begin{table}[t]
    \centering
    \small
    \setlength{\tabcolsep}{4pt}
    \renewcommand{\arraystretch}{1.00}
    \caption{Comparison of representative Real-ISR families. Compared with other methods, our SANA-SR simultaneously achieves one-step inference without training from scratch, $32{\times}$ latent compression, linear-complexity attention, and structured pruning.
    }
    \vspace{-2mm}
    \label{tab:paradigm_comparison}
    \resizebox{\textwidth}{!}{%
    \begin{tabular}{lcccccccccc}
    \toprule
    \textbf{Backbone} & \makecell[c]{\textbf{Representative} \\ \textbf{work}} & \makecell[c]{\textbf{One} \\ \textbf{step} } &  \makecell[c]{\textbf{Train} \\ \textbf{from scratch}}
    & \makecell[c]{\textbf{Compress.} \\ \textbf{scale}} & \makecell[c]{\textbf{\#tokens} \\ $N$}& \makecell[c]{\textbf{Attn.}\\ \textbf{complexity}} & \makecell[c]{\textbf{\#Params} \\ $\bf{{<}0.5B}$ } & \textbf{Prune} & \makecell[c]{\textbf{Time} \\ $\bf{{<}20ms}$ } \\
    \midrule
    \multirow{2}{*}{SD \citep{rombach2022high}}          & OSEDiff~\citep{wu2024osediff}    & \cmark   & \xmark & $8\times$    & $\sim$4K  & $\mathcal{O}(N^2)$ & \xmark  & --         & \xmark       \\
        & AdcSR~\citep{chen2025adcsr}      & \cmark   & \xmark & $8\times$    & $\sim$4K  & $\mathcal{O}(N^2)$ & \cmark             & \cmark & \xmark      \\
    FLUX \citep{flux2024,labs2025flux1kontextflowmatching}  & FluxSR~\citep{li2025one}         & \cmark   & \xmark & $8\times$    & $\sim$4K  & $\mathcal{O}(N^2)$ & \xmark  & --         & \xmark \\
    \multirow{2}{*}{LinearAttn \citep{katharopoulos2020transformers}}  & LinearSR~\citep{li2026linearsr}  & \xmark  & \cmark & $32\times$   & 256       & $\mathcal{O}(N)$ & \xmark         & --         & \xmark     \\
    {}      & \textbf{SANA-SR}                 & \cmark & \xmark & $32\times$ & 256 & $\mathcal{O}(N)$ & \cmark & \cmark & \cmark \\
    \bottomrule
    \end{tabular}%
    }
    \vspace{-5mm}
\end{table}

\textbf{\cao{Dense token representation and interaction in Real-ISR.}}
Earlier SR approaches mainly rely on feed-forward regression networks~\citep{dong2014learning, lim2017enhanced, zhang2018image, wang2018esrgan, liang2021swinir, chen2023activating} with hand-crafted blind degradation models~\citep{zhang2021designing, wang2021real}. In contrast, a major recent trend is to solve Real-ISR by adapting large pretrained generative models. Early diffusion-based methods such as StableSR \citep{wang2024stablesr} uses pretrained text-to-image diffusion models to improve perceptual realism. Subsequent work strengthened this paradigm by introducing richer conditioning or larger restoration backbones: SeeSR \citep{wu2024seesr} uses degradation-aware semantic prompting to better preserve image semantics; SUPIR \citep{yu2024supir} pushes restoration quality further through model scaling and prompt-guided restoration; DiffBIR decouples degradation removal from information regeneration \citep{lin2024diffbir}; and DreamClear and DiT4SR move this line toward stronger DiT-based restoration models with more expressive degradation modeling and low-resolution guidance injection \citep{ai2024dreamclear,duan2025dit4sr}. 

\textbf{\cao{One-step diffusion or flow-matching based Real-ISR.}} 
A second thread focuses on reducing inference latency to one or a few sampling steps. SinSR \citep{wang2024sinsr} shows that a single-step SR model can be obtained from a deterministic diffusion mapping. OSEDiff takes the low-quality image itself as the diffusion starting point and regularizes one-step prediction with a latent variational score-distillation objective \citep{wu2024osediff}. TSD-SR further strengthens one-step restoration with a target-score-matching formulation, while DoSSR and InvSR explore efficiency from different directions, namely domain-shift diffusion and flexible few-step inversion, respectively \citep{dong2025tsdsr,cui2024dossr,yue2025invsr}. More recent methods such as AdcSR, PiSA-SR, and D$^3$SR investigate controllability, compression, and stronger single-step training objectives \citep{chen2025adcsr,sun2025pisasr,li2025d3sr}. Beyond diffusion, flow-matching-based frameworks such as FLUX-SR cast Real-ISR as a one-step flow trajectory \citep{li2025one}. These works substantially reduce the number of function evaluations, but typically retain the dense high-resolution latent representations and quadratic-cost token interactions of standard diffusion backbones, so that the per-step cost is still dominated by token-level computation. 
In contrast, we target the per-step cost itself by combining a compact latent representation with linear-complexity token interaction, which is complementary to the sampling-step reductions above.

\textbf{Architecture-Level Efficiency in Generative SR.} Beyond reducing the number of diffusion steps, a newer line of work asks whether the backbone itself can be redesigned for HR efficiency. SANA shows that linear-attention diffusion transformers combined with deep latent compression can support HR generation at much smaller computational cost than standard quadratic-attention backbones \citep{xie2025sana}. Most closely related to our work, LinearSR demonstrates that linear attention can be stabilized for photorealistic SR through a carefully engineered training recipe involving ESGF, SNR-MoE, and tag-style guidance \citep{li2026linearsr}. Very recent deployment-oriented efforts, such as Q-DiT4SR, further explore post-training efficiency for DiT-based restoration via quantization \citep{zhang2026qdit4sr}. Relative to these works, our contribution is different in scope: we do not introduce a new multi-step LinearDiT SR architecture from scratch, nor do we focus solely on post hoc acceleration of a large DiT restorer. Instead, we repurpose a \emph{compact pretrained LinearDiT prior} into a one-step SR model and then compress it further with prompt-aware structured pruning, complementing the long line of network pruning techniques~\citep{lecun1989optimal, han2015learning, liu2017learning, frankle2018lottery, hoefler2021sparsity, fang2023depgraph, ma2023llm, fang2025tinyfusion}.  This places our method at the intersection of fast diffusion SR and architecture-level efficiency.

\section{Proposed Method}
\label{sec:method}
\vspace{-2mm}
\cao{The key challenge is that existing generative Real-ISR methods typically operate on dense latent tokens and rely on costly quadratic token interactions, making high-resolution restoration inefficient. To address this issue, we design a efficient one-step restoration method, called SANA-SR, as illustrated in \ref{fig:framework}.
Given a low-quality (LQ) input, a targeted prompt extractor first produces semantic tags, which are concatenated with a fixed quality suffix and passed through the frozen tokenizer-text encoder to form the refined text condition. Meanwhile, the deep-compression VAE encoder maps both the LQ input and the high-quality (HQ) target into a compact $32\times$ latent space, yielding $z_L$ and $z_H$. Conditioned on the refined prompt, a LoRA-adapted one-step LinearDiT predicts a latent residual and updates $z_L$ to the restored latent $\hat z$, which is then decoded by the frozen decoder to produce the SR output. During training, we supervise the model with pixel and perceptual losses, and further regularize one-step adaptation with matched-noise frozen-prior alignment and adapter-on/off consistency. After training, the adapted backbone is further compressed by prompt-aware structured pruning, which accumulates block saliency under the same task loss and calibration tuples, and selects a subset of transformer blocks under a deployment budget for efficient inference.
}

\begin{figure}[t]
	\centering
	\includegraphics[width=\linewidth]{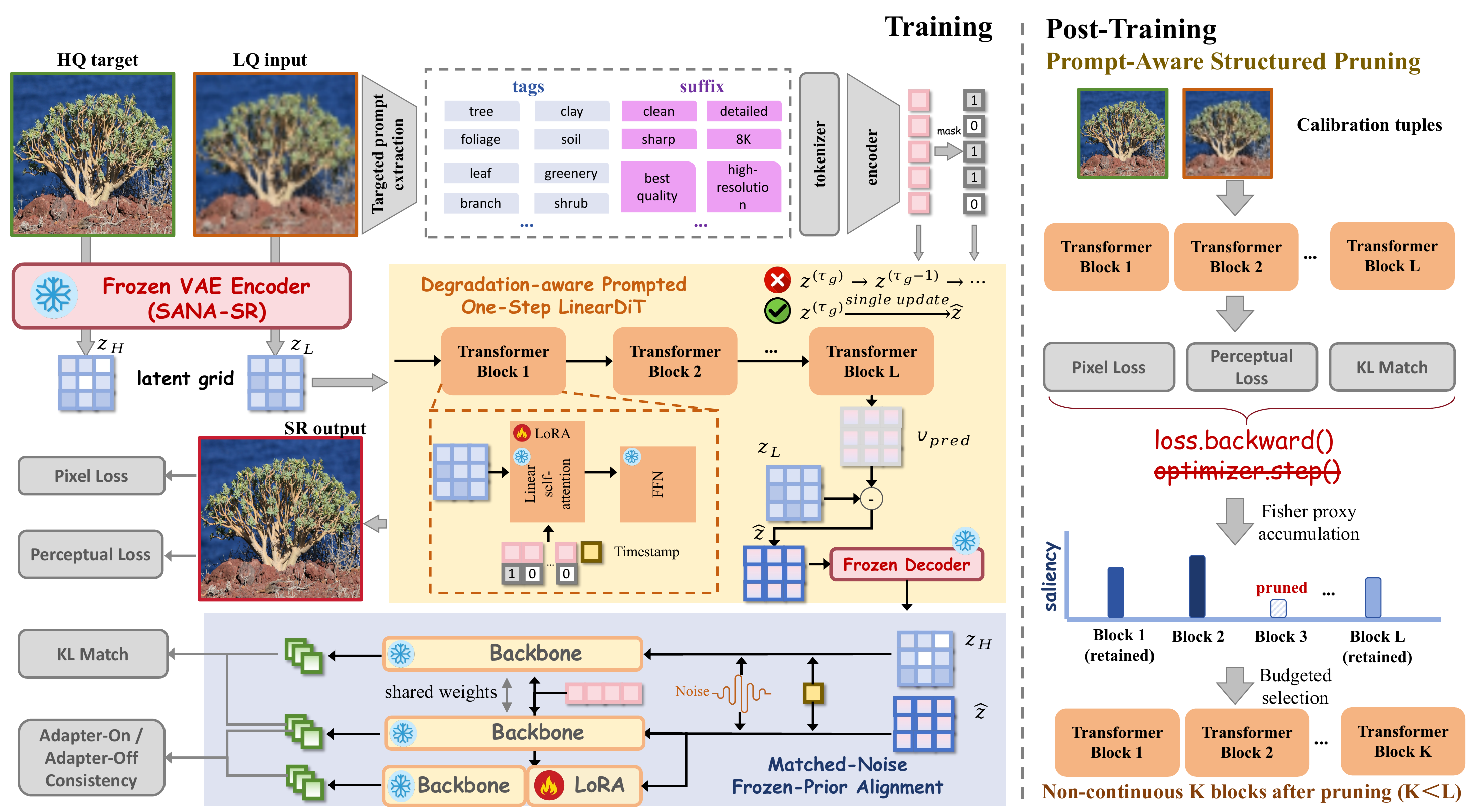}
    \vspace{-5mm}
    \caption{\cao{\textbf{Overview of SANA-SR.} Given an LQ input, SANA-SR first maps the image into a compact latent space with a frozen deep-compression VAE, then restores the latent with a prompt-conditioned one-step LinearDiT adapted by LoRA. Training is regularized by frozen-prior alignment and adapter consistency, and the final model is further compressed by prompt-aware structured pruning for efficient deployment.}}
	\label{fig:sanasr_pipeline}
    \vspace{-2mm}
\end{figure}

\subsection{Degradation-Stable Compact Latent Representation}
\label{sec:method:dcae}

Real-ISR inputs are corrupted by an unknown mixture of blur, noise, codec artifacts, and sensor non-idealities. 
\cao{Although such degradations severely distort local pixel observations, they often preserve the underlying scene layout and semantic structure. This suggests that restoration may be better posed in a compact latent space than in dense pixel-aligned latent grids. In particular, if the latent representation is more stable than the pixel space under common degradations, then Real-ISR can be reduced to a smaller latent correction problem that is more suitable for one-step restoration.}

\cao{
To this end, we adopt the pretrained SANA deep-compression autoencoder (DC-AE)~\citep{xie2025sana} as a frozen visual front-end. Although the DC-AE is not trained specifically for degradation modeling, we empirically observe that its compressed latent representation is substantially less sensitive to common real-world degradations than the pixel space. This property allows us to perform restoration in a compact latent space while preserving restoration-relevant structure and texture cues.}
\cao{For a low-quality input $x_L\in\mathbb{R}^{3{\times}H{\times}W}$
 and its high-quality target $x_H\in\mathbb{R}^{3{\times}H{\times}W}$
, we define}
\begin{equation}
z_L = E(x_L),\qquad z_H = E(x_H),\qquad
z_L,z_H \in \mathbb{R}^{C\times h\times w},\;\; h=H/32,\; w=W/32,
\label{eq:main_vae}
\end{equation}
where $E$ is the frozen deep compression encoder.
For a $512{\times}512$ image this yields $N{=}h{\cdot}w=256$ latent tokens.
\cao{Our method has fewer tokens than most existing diffusion-based restorers ~\citep{wang2024stablesr,lin2024diffbir,wu2024seesr,wu2024osediff,chen2025adcsr,dong2025tsdsr} which operate on dense $4\times$ or $8\times$ VAE latents, leading to high per-step computation.}
The encoder and decoder are kept frozen throughout training, inference, and pruning calibration.



\subsection{Degradation-aware Prompted One-Step LinearDiT}
\label{sec:method:onestepdit}
\cao{Once the restoration problem is transferred to a compact latent space, the next question is how to effectively adapt a generative prior to this one-step SR setting. Rather than training a new restoration transformer from scratch, we adapt a pretrained SANA LinearDiT \citep{xie2025sana} prior with lightweight LoRA updates. To make the one-step prediction consistent with clean image generation rather than degraded input appearance, we further condition the model on a degradation-aware refined prompt.}


\textbf{Degradation-aware prompt refinement.}
\cao{A caption extracted directly from the low-quality input $x_L$ often contains degradation-related words such as “blurry”, “noisy”, or “low-quality”. Conditioning the generative prior on such tokens is counter-productive for restoration, since it biases prediction toward degraded outputs. To avoid this issue, we construct a refined prompt from three components:}
(i) a tag-based extractor $\Pi(\cdot)$, that emits content tags from $x_L$ (e.g.\ \emph{tree}, \emph{foliage}, \emph{leaf}, \emph{branch}); (ii) a fixed quality template $p_{\mathrm{tpl}}=$~\emph{``clean, sharp, best quality, detailed, 8K, high-resolution''} that biases the prior toward the clean regime; and (iii) an attention mask $m$ that suppresses any residual degradation-related token. Letting $\mathcal{T}(\cdot)$ denote the frozen SANA tokenizer-text encoder, the corrected prompt embedding is
\begin{equation}
(c,m)=\mathcal{T}\!\left(\Pi(x_L) \oplus p_{\mathrm{tpl}}\right),
\qquad
c\in\mathbb{R}^{T\times d_t},\;\; m\in\{0,1\}^{T},
\label{eq:main_prompt}
\end{equation}
where $\oplus$ denotes string concatenation, $c$ is the resulting text-embedding sequence and $m$ its attention mask (with $m_i{=}0$ marking padding/suppressed positions), $T$ is the token length, and $d_t$ is the text-embedding dimension.  

\textbf{One-step LinearDiT update.}
\cao{Next, we use a one-step LinearDiT to predict the restored latent. The LinearDiT backbone consists of $L$ transformer blocks, each containing linear self-attention, cross-attention to the text embedding, and an FFN. We first flatten $z_L$ into a token sequence of length $N{=}hw$. In each block, the queries $Q$, keys $K$, and values $V$ are three learned linear projections of this latent token sequence, so that}
\begin{equation}
\operatorname{LA}(Q,K,V)
= \frac{\phi(Q)\big(\phi(K)^{\!\top} V\big)}{\phi(Q)\big(\phi(K)^{\!\top} \mathbf{1}\big)+\varepsilon_{\mathrm{att}}},
\label{eq:main_la}
\end{equation}
where $\phi(\cdot)$ is a non-negative feature map and $\varepsilon_{\mathrm{att}}{>}0$ is a stabilizer. 
\cao{The refined prompt enters through cross-attention, where keys and values are projected from $c$, and the mask $m$ suppresses padding or filtered token positions.}

Let $f_0$ denote the frozen pretrained LinearDiT and $f_\theta$ its LoRA-adapted version, where LoRA~\citep{hu2022lora} is inserted into the attention (self- and cross-attention) and FFN projections only. At a fixed generation timestep $\tau_g$, $f_\theta$ predicts a latent residual from $z_L$ and applies one latent update:
\begin{equation}
\hat z = z_L - \sigma_{\tau_g}\, f_\theta(z_L,\tau_g,c,m),
\qquad
\hat x = D(\hat z),
\label{eq:main_onestep}
\end{equation}

where $\sigma_{\tau_g}{>}0$ is the scheduler coefficient at $\tau_g$, $\hat z$ is the restored latent, and $\hat x$ is the SR output from the frozen VAE decoder $D$. Eq.~\eqref{eq:main_onestep} is also the inference path of SANA-SR: at test time only the solid path in Fig.~\ref{fig:sanasr_pipeline} is executed, without the alignment branch and without any iterative diffusion trajectory.

\vspace{-3.5mm}
\paragraph{Linear complexity.}
\cao{With $32{\times}$ latent compression, a $512{\times}512$ input yields only $N{=}256$ spatial tokens. Combined with linear attention, the resulting per-image complexity is $\mathcal{O}(N)$, which is substantially lower than quadratic-attention SR backbones operating on dense latent grids. More discussions are in Appendix~\ref{app:lineardit}.}

\subsection{Frozen-Prior Alignment and Adapter Consistency}
\label{sec:method:alignment}

Image-space supervision alone does not constrain whether the one-step latent update stays compatible with the pretrained prior, which can drift in our compact-model setting. We therefore reuse the frozen SANA $f_0$ (the same prior we adapt with LoRA) as a reference probe: if $\hat z$ is close to $z_H$, then under matched perturbation and text condition the frozen prior should respond similarly.

We sample a training timestep $t$ and a Gaussian perturbation $\epsilon \sim \mathcal{N}(0,I)$, and form the perturbed restored and reference latents
$\tilde z_{\hat z} {=} \alpha_t \hat z {+} \sigma_t \epsilon,
\tilde z_H {=} \alpha_t z_H {+} \sigma_t \epsilon$,
where $(\alpha_t,\sigma_t)$ are the scheduler coefficients at timestep $t$. We then evaluate the frozen pretrained backbone on both perturbed latents under the same text condition,
$
q_{\hat z}{=}f_0(\tilde z_{\hat z},t,c,m),
q_H{=}f_0(\tilde z_H,t,c,m),
$
and, denoting by $\mu_{b,c}(q)$ and $s_{b,c}(q)$ the spatial mean and variance of channel $c$ in sample $b$ of a response tensor $q$, we align the two responses by matching their channel-wise Gaussian summaries:
\begin{equation}
\mathcal{L}_{\mathrm{align}}
=
\frac{1}{2BC}
\sum_{b=1}^{B}\sum_{c=1}^{C}
\left[
\log \frac{s_{b,c}(q_H)+\varepsilon}{s_{b,c}(q_{\hat z})+\varepsilon}
+
\frac{
s_{b,c}(q_{\hat z})+
\big(\mu_{b,c}(q_{\hat z})-\mu_{b,c}(q_H)\big)^2
}{
s_{b,c}(q_H)+\varepsilon
}
-1
\right],
\label{eq:main_align}
\end{equation}
where $B$ is the batch size, $C$ is the number of response channels, and $\varepsilon>0$ is a numerical stabilizer.

To keep the learned adapter close to the frozen prior, we further compare the adapter-on and adapter-off responses on the same perturbed restored latent:
\begin{equation}
\mathcal{L}_{\mathrm{cons}}
=
{1}/{|\Omega|}
\left\|
f_\theta(\tilde z_{\hat z},t,c,m)-f_0(\tilde z_{\hat z},t,c,m)
\right\|_2^2,
\label{eq:main_cons}
\end{equation}
where $|\Omega|$ is the number of scalar elements in the response tensor.

The image-space reconstruction term combines pixel and perceptual losses:
\begin{equation}
\mathcal{L}_{\mathrm{rec}}
=
\lambda_{2}\frac{1}{3HW}\|\hat x-x_H\|_2^2
+
\lambda_{p}\,\operatorname{LPIPS}(\hat x,x_H),
\label{eq:main_rec}
\end{equation}
where 
$\lambda_2,\lambda_p>0$ weight the pixel and perceptual terms. The final training objective is
\begin{equation}
\mathcal{L}
=
\mathcal{L}_{\mathrm{rec}}
+
\lambda_{a}\mathcal{L}_{\mathrm{align}}
+
\lambda_{c}\mathcal{L}_{\mathrm{cons}},
\label{eq:main_total}
\end{equation}
where $\lambda_a,\lambda_c>0$ control the strength of frozen-prior alignment and adapter consistency. 
Only the LoRA parameters in $f_\theta$ are optimized; the VAE, text encoder, and pretrained SANA remain frozen.

\subsection{Prompt-Aware Structured Pruning}
\label{sec:method:pruning}
\vspace{-2mm}
\cao{After LoRA adaptation, we further compress the deployed model through structured block pruning. Since SANA-SR is text-conditioned, pruning should preserve prompt-conditioned restoration behavior rather than generic denoising ability.}

We first merge the learned LoRA adapters into the LinearDiT backbone and obtain a dense model $\bar f_\theta$. Let the LinearDiT contain $L$ transformer blocks $\{\mathcal{B}_\ell\}_{\ell=1}^{L}$. For pruning calibration, we reuse the SR task loss without the adapter-consistency term,
\(
\mathcal{L}_{\mathrm{cal}}
=
\mathcal{L}_{\mathrm{rec}}
+
\lambda_a \mathcal{L}_{\mathrm{align}}.
\)
For every scalar parameter $w_i$ in the merged backbone we estimate a diagonal curvature proxy from $K$ calibration steps,
\begin{equation}
F_i
=
\frac{1}{K}
\sum\nolimits_{k=1}^{K}
\omega(t_k)
\left(
\left.\partial \mathcal{L}_{\mathrm{cal}}^{(k)} \right/ {\partial w_i}
\right)^2,
\label{eq:main_fisher}
\end{equation}
where $t_k$ is the calibration timestep at iteration $k$ and $\omega(t_k)$ is a timestep-dependent weight. We then define the saliency of block $\mathcal{B}_\ell$ by the reciprocal-curvature-weighted score
\begin{equation}
S_\ell
=
\sum\nolimits_{w_i \in \mathcal{B}_\ell}
\left.{w_i^2}\right/({F_i+\varepsilon_p}),
\label{eq:main_score}
\end{equation}
where $\varepsilon_p{>}0$ is a stabilizer. 
\cao{A larger $S_\ell$ indicates a larger estimated loss increase from removing $\mathcal{B}_\ell$.}

Let $P_\ell$ be the parameter count of block $\mathcal{B}_\ell$, $P_{\mathrm{fix}}$ the non-prunable parameters, and $P_\star$ the target deployment budget. We select the kept block index set $\mathcal{K}$ by
\begin{equation}
\max_{\mathcal{K}\subseteq\{1,\dots,L\}}
\sum\nolimits_{\ell\in\mathcal{K}} S_\ell
\quad
\text{s.t.}
\quad
P_{\mathrm{fix}}+\sum\nolimits_{\ell\in\mathcal{K}}P_\ell \le P_\star,
\qquad
\{1,L\}\subseteq\mathcal{K},
\label{eq:main_budget}
\end{equation}
so that the first and last transformer blocks are retained for stability. After selecting $\mathcal{K}$, we remove the remaining blocks and obtain a pruned one-step model with the same inference form as Eq.~\eqref{eq:main_onestep}.


\vspace{-2mm}
\section{Experimental Results}
\vspace{-2mm}
\subsection{Experimental Setup}
\vspace{-2mm}
\label{sec:exp_setup}

\textbf{Datasets.}
SANA-SR is trained on a mixed dataset 
combining the DIV2K~\citep{agustsson2017ntire}, Flickr2K~\citep{lim2017enhanced}, LSDIR~\citep{li2023lsdir}, and FFHQ~\citep{karras2019style}, with the degradations synthesized by Real-ESRGAN pipeline~\citep{wang2021real};
the released pipeline also supports optionally augmenting the HQ pool with RealSR~\citep{cai2019toward} training images. 

\vspace{-1mm}
\textbf{Evaluation metrics.}
We use PSNR, SSIM, MANIQA, MUSIQ, and CLIPIQA to measure distortion fidelity and no-reference perceptual quality. PSNR and SSIM are computed on the Y channel of the YCbCr space, while the no-reference metrics are computed on RGB outputs. Unless otherwise stated, results are reported under a full-image protocol; for methods whose official reports use cropped evaluation, we additionally provide a $256\times256$ center-crop patch protocol for fair comparison.

\vspace{-1mm}
\textbf{Compared methods.}
We compare SANA-SR against two groups: (i) Prior-based multi-step or flexible-step restorers, including StableSR~\citep{wang2024stablesr}, DiffBIR~\citep{lin2024diffbir}, SeeSR~\citep{wu2024seesr}, SUPIR~\citep{yu2024supir}, DreamClear~\citep{ai2024dreamclear}, InvSR~\citep{yue2025invsr}, and LinearSR~\citep{li2026linearsr}. 
(ii) Efficient or one-step methods, including SinSR~\citep{wang2024sinsr}, OSEDiff~\citep{wu2024osediff}, AdcSR~\citep{chen2025adcsr}, and TSD-SR~\citep{dong2025tsdsr}. 
For fairness, each baseline is evaluated with its official checkpoint and default inference setting whenever available, while the test set, upscaling factor, image protocol (full-image or patch), and metric implementation are kept identical across methods.

\vspace{-1mm}
\textbf{Implementation details.}
Experiments are conducted on $8\times$ NVIDIA RTX 4090 with FP16 training and $100$K AdamW steps (learning rate $5\times 10^{-5}$, batch size 4, $512\times 512$ random crops, $4\times$ SR). We freeze the VAE and text encoder and optimize only LoRA layers (rank 64, scale 64) inserted into the LinearDiT. Loss weights: $\lambda_p{=}2$, $\lambda_2{=}\lambda_a{=}\lambda_c{=}1$. Full hyperparameter list is in Appendix~\ref{app:exp_details}.
\cao{We instantiate the extractor $\Pi$ with DAPE \citep{wu2024seesr}.
More details are put in Appendix~\ref{app:prompt}.}

\vspace{-2mm}
\subsection{Main Results}
\vspace{-2mm}

\begin{table*}[t]
\centering
\caption{Quantitative comparison on DIV2K-Val with \textcolor{MyRed}{best} and \textcolor{MyBlue}{second-best} results.}
\label{tab:main_div2k}
\resizebox{\textwidth}{!}{%
\renewcommand{\arraystretch}{1.08}
\begin{tabular}{l*{9}{c}}
\toprule
Method & Venue & PSNR$\uparrow$ & SSIM$\uparrow$ & MANIQA$\uparrow$ & MUSIQ$\uparrow$ & CLIPIQA$\uparrow$ & LPIPS$\downarrow$ & DISTS$\downarrow$ & NIQE$\downarrow$ \\
\midrule
\rowcolor{gray!15} \multicolumn{10}{l}{\emph{Prior-based multi-step / flexible-step methods}} \\
StableSR~\citep{wang2024stablesr}   & IJCV'24 & 23.26 & 0.5726 & 0.6192 & 65.92 & 0.6771 & 0.3113 & 0.2048 & 4.7581\\
DiffBIR~\citep{lin2024diffbir}      & ECCV'24 & 23.64 & 0.5647 & 0.6210 & 65.81 & 0.670  & 0.3524 & 0.2128 & 4.7042\\
SeeSR~\citep{wu2024seesr}           & CVPR'24 & 23.68 & 0.6043 & 0.6240 & 68.67 & 0.6936 & 0.3194 & 0.1968 & 4.8102\\
PASD~\citep{yang2024pixel}          & ECCV'24 & 23.14 & 0.5505 & \second{0.6483} & 68.95 & 0.6788 & 0.3571 & 0.2207 & 4.3617\\
ResShift~\citep{yue2023resshift}    & NeurIPS'23 & \second{24.65} & 0.6181 & 0.5454 & 61.09 & 0.6071 & 0.3349 & 0.2213 & 6.8212\\
SUPIR~\citep{yu2024supir}           & CVPR'24 & 22.13 & 0.5279 & 0.5903 & 63.86 & 0.7146 & 0.3919 & 0.2312 & 5.6767 \\
DreamClear~\citep{ai2024dreamclear} & NeurIPS'24 & 22.03 & 0.5415 & 0.6320 & 68.44 & 0.6725 & 0.3189 & \best{0.1719} & 5.3126 \\
InvSR~\citep{yue2025invsr}          & CVPR'25 & 24.32 & 0.6309 & 0.4291 &  69.46 & 0.675 & 0.2821 & 0.2214 & 4.3428\\
LinearSR~\citep{li2026linearsr}     & ICLR'26 & 24.53 & \second{0.6349} & 0.4732 &  70.14 & 0.683 & 0.2776 & 0.2307 & 4.4854\\
\midrule
\rowcolor{gray!15} \multicolumn{10}{l}{\emph{Efficient / one-step methods}} \\
SinSR~\citep{wang2024sinsr}         & CVPR'24 & 24.41 & 0.6018 & 0.5386 & 62.82 & 0.6471 & 0.3240 & 0.2066 & 6.0159 \\
OSEDiff~\citep{wu2024osediff}       & NeurIPS'24 & 23.72 & 0.6109 & 0.6131 & 67.96 & 0.6681 & 0.2941 & 0.1976 & 4.7097 \\
S3Diff~\citep{zhang2024degradation} & ArXiv'24   & 23.40 & 0.5953 & 0.5538 & 68.21 & 0.7007 & \best{0.2571} & 0.1930 & 4.7391 \\
AddSR~\citep{tai2026addsr}          & Pattern Recognit'26 & 22.16 & 0.6280 & \best{0.6596} & \second{70.99} & \best{0.7593} & 0.4053 & 0.2360 & 5.2584  \\
D$^3$SR~\citep{li2025d3sr}    & NeurIPS'25 & 21.95 & 0.6037 & 0.6271 & 70.35 & 0.6828 & 0.3076 & 0.1913 & 4.4580 \\
FLUX-SR~\citep{li2025one}           & ICML'25 & 22.83 & 0.6177 & 0.6401 & 69.92 & 0.7030 & 0.2717 & 0.1834 & 5.0626 \\
AdcSR~\citep{chen2025adcsr}         & CVPR'25 & 23.78 & 0.6023 & 0.6300 & 69.66 & 0.6765 & 0.3073 & 0.2007 & 4.6631\\
TSD-SR~\citep{dong2025tsdsr}        & CVPR'25 & 23.87 & 0.5808 & 0.6192 & 70.69 & \second{0.7416} & \second{0.2673} & 0.1821 & \second{4.3244} \\
\midrule
\textbf{Ours} & -- & \best{24.92} & \best{0.6429} & 0.6412 & \best{71.32} & 0.7005 & 0.3075 & \second{0.1817} & \best{4.1354} \\
\bottomrule
\end{tabular}%
\vspace{-4mm}
}
\end{table*}

\begin{table*}[t]
\vspace{-5mm}
\centering
\caption{Quantitative comparison on RealSR with \textcolor{MyRed}{best} and \textcolor{MyBlue}{second-best} results.}
\label{tab:main_realsr}
\resizebox{\textwidth}{!}{%
\renewcommand{\arraystretch}{1.08}
\begin{tabular}{l*{9}{c}}
\toprule
Method & Venue & PSNR$\uparrow$ & SSIM$\uparrow$ & MANIQA$\uparrow$ & MUSIQ$\uparrow$ & CLIPIQA$\uparrow$ & LPIPS$\downarrow$ & DISTS$\downarrow$ & NIQE$\downarrow$ \\
\midrule
\rowcolor{gray!15} \multicolumn{10}{l}{\emph{Prior-based multi-step / flexible-step methods}} \\
StableSR~\citep{wang2024stablesr}   & IJCV'24 & 24.70 & 0.7085 & 0.6221 & 65.78 & 0.6178 & 0.3018 & 0.2288 & 5.9122 \\
DiffBIR~\citep{lin2024diffbir}      & ECCV'24 & 24.75 & 0.6567 & 0.6246 & 64.98 & 0.6463 & 0.3636 & 0.2312 & 5.5346 \\
SeeSR~\citep{wu2024seesr}           & CVPR'24 & 25.18 & 0.7216 & 0.6442  & 69.77  & 0.6612 & 0.3009 & 0.2223 & 5.4081\\
PASD~\citep{yang2024pixel}          & ECCV'24 & 25.21 & 0.6798 & \second{0.6487} & 68.75 & 0.6620 & 0.3380 & 0.2260 & 5.4137\\
ResShift~\citep{yue2023resshift}    & NeurIPS'23 & \best{26.31} & \second{0.7421} & 0.5285 & 58.43 & 0.5444 & 0.3460 & 0.2498 & 7.2635\\
SUPIR~\citep{yu2024supir}           & CVPR'24 & 23.65 & 0.6620 & 0.5780 & 62.09 & 0.6707 & 0.3541 & 0.2488 & 6.1099\\
DreamClear~\citep{ai2024dreamclear} & NeurIPS'24 & 22.56 & 0.6548 & 0.5384 & 65.21 & 0.6895 & 0.3684 & 0.2352 & 5.7381  \\
InvSR~\citep{yue2025invsr}          & CVPR'25 & 24.50 & 0.7260 & 0.4456 & 69.67 & 0.6918 & 0.2978 & 0.2492 & 5.2189\\
LinearSR~\citep{li2026linearsr}     & ICLR'26 & 23.84 & 0.6848 & 0.6108 & 69.39 & 0.6731 & 0.3128 & 0.2935 & 5.8509 \\
\midrule
\rowcolor{gray!15} \multicolumn{10}{l}{\emph{Efficient / one-step methods}} \\
SinSR~\citep{wang2024sinsr}         & CVPR'24 & \second{25.98} & 0.7347 & 0.5385 & 60.80 & 0.6122 & 0.3188 & 0.2353 & 6.2872 \\
OSEDiff~\citep{wu2024osediff}       & NeurIPS'24 & 25.15 & 0.7341 & 0.6326 & 69.09 & 0.6693 & 0.2921 & 0.2128 & 5.6476\\
S3Diff~\citep{zhang2024degradation} & ArXiv'24   & 25.03 & 0.7321 & 0.6263 & 67.89 & 0.6722 & \best{0.2699} & \second{0.1996} & 5.3311 \\
AddSR~\citep{tai2026addsr}          & Pattern Recognit'26 & 23.33 & 0.6400 & \best{0.6826} & \best{71.49} & 0.7225 & 0.3925 & 0.2626 & 5.8959  \\
D$^3$SR~\citep{li2025d3sr}    & NeurIPS'25 & 24.54 & 0.7270 & 0.6382 & 68.69 & 0.6711 & 0.3050 & 0.2106 & \second{5.0960}  \\
FLUX-SR~\citep{li2025one}           & ICML'25 & 24.83 & 0.7375 & 0.6511 & 70.08 & \second{0.7381} & 0.3141 & 0.2264 & 5.2097\\
AdcSR~\citep{chen2025adcsr}         & CVPR'25 & 25.31 & 0.7238 & 0.6372 & 70.31 & 0.7363 & 0.2997 & 0.2162 & 5.3152 \\
TSD-SR~\citep{dong2025tsdsr}        & CVPR'25 & 24.81 & 0.7172 & 0.6347 & 70.49 & 0.7160 & \second{0.2743} & 0.2104 & 5.1298 \\
\midrule
\textbf{Ours} & -- & 25.96 & \best{0.7468} & 0.5932 & \second{70.73} & \best{0.7663} & 0.2917 & \best{0.1964} & \best{5.0897}\\
\bottomrule
\end{tabular}%
}
\vspace{-5mm}
\end{table*}

\begin{table*}[htp]
\vspace{-5mm}
\centering
\caption{Quantitative comparison on DRealSR with \textcolor{MyRed}{best} and \textcolor{MyBlue}{second-best} results.}
\label{tab:main_drealsr}
\resizebox{\textwidth}{!}{%
\renewcommand{\arraystretch}{1.08}
\begin{tabular}{l*{9}{c}}
\toprule
Method & Venue & PSNR$\uparrow$ & SSIM$\uparrow$ & MANIQA$\uparrow$ & MUSIQ$\uparrow$ & CLIPIQA$\uparrow$ & LPIPS$\downarrow$ & DISTS$\downarrow$ & NIQE$\downarrow$\\
\midrule
\rowcolor{gray!15} \multicolumn{10}{l}{\emph{Prior-based multi-step / flexible-step methods}} \\
StableSR~\citep{wang2024stablesr}   & IJCV'24 & 28.03 & 0.7536 & 0.5601 & 58.51 & 0.6356 & 0.3284 & 0.2269 & 6.5239\\
DiffBIR~\citep{lin2024diffbir}      & ECCV'24 & 26.71 & 0.6571 & 0.5930 & 61.07 & 0.6395 & 0.4557 & 0.2748 & 6.3124\\
SeeSR~\citep{wu2024seesr}           & CVPR'24 & 28.17 & 0.7691 & 0.6042 & 64.93  & 0.6804 & 0.3189 & 0.2315 & 6.3967 \\
PASD ~\citep{yang2024pixel}         & ECCV'24 & 27.36 & 0.7073 & \second{0.6169} & 64.87 & 0.6808 & 0.3760 & 0.2531 & 5.5474\\
ResShift~\citep{yue2023resshift}    & NeurIPS'23 & \best{28.46} & 0.7673 & 0.4586 & 50.60 & 0.5342 & 0.4006 & 0.2656 & 8.1249\\
SUPIR~\citep{yu2024supir}           & CVPR'24 & 25.09 & 0.6460 & 0.5471 & 58.79 & 0.6749 & 0.4243 & 0.2795 & 7.3918\\
DreamClear~\citep{ai2024dreamclear} & NeurIPS'24 & 24.48 & 0.6508 & 0.4465 & 65.83 & 0.6620 & 0.3972 & 0.2445 & \second{5.1326}\\
InvSR~\citep{yue2025invsr}          & CVPR'25 & 27.63 & 0.7962 & 0.4610 & 67.46 & 0.6918 & \best{0.2897} & 0.2373 & 6.3219\\
LinearSR~\citep{li2026linearsr}     & ICLR'26 & 26.91 & 0.7190 & 0.5812 & \second{69.22} & 0.7132 & 0.3584 & 0.3003 & 6.9654\\
\midrule
\rowcolor{gray!15} \multicolumn{10}{l}{\emph{Efficient / one-step methods}} \\
SinSR~\citep{wang2024sinsr}         & CVPR'24 & \second{28.36} & 0.7515 & 0.4884 & 55.33 & 0.6383 & 0.3665 & 0.2485 & 6.9907\\
OSEDiff~\citep{wu2024osediff}       & NeurIPS'24 & 27.92 & 0.7835 & 0.5899 & 64.65 & 0.6963 & 0.2968 & 0.2165 & 6.4902 \\
S3Diff~\citep{zhang2024degradation} & ArXiv'24   & 27.39 & 0.7469 & 0.5723 & 64.16 & 0.7156 & 0.3129 & \second{0.2108} & 6.1700 \\
AddSR~\citep{tai2026addsr}          & Pattern Recognit'26 & 26.72 & 0.7124 & \best{0.6257} & 66.33 & \second{0.7226} & 0.3982 & 0.2711 & 7.6689  \\
D$^3$SR~\citep{li2025d3sr}    & NeurIPS'25 & 26.98 & 0.7135 & 0.5964 & 67.28 & 0.7086 & 0.3083 & 0.2226 & 5.5233 \\
FLUX-SR~\citep{li2025one}           & ICML'25 & 27.29 & \second{0.7963} & 0.5992 & 68.79 & 0.6730 & \second{0.2902} & 0.2290 & 5.9302 \\
AdcSR~\citep{chen2025adcsr}         & CVPR'25 & 28.10 & 0.7726 & 0.6045 & 66.26 & 0.7049 & 0.3046 & 0.2200 & 6.4500\\
TSD-SR~\citep{dong2025tsdsr}        & CVPR'25 & 27.77 & 0.7559 & 0.5874 & 66.62 & \best{0.7344} & 0.2967 & 0.2136 & 5.9131\\
\midrule
\textbf{Ours} & -- & 27.99 & \best{0.8001} & 0.6073 & \best{70.69} & 0.7179 & 0.3275 & \best{0.2033} & \best{5.1311}\\
\bottomrule
\end{tabular}%
}
\vspace{-5mm}
\end{table*}

\noindent\textbf{DIV2K-Val.} Table~\ref{tab:main_div2k} shows that SANA-SR achieves the best PSNR, SSIM, and MUSIQ on DIV2K-Val, while ranking second on DISTS and NIQE. This indicates that our compact one-step model preserves strong structural fidelity under controlled synthetic degradations while remaining highly competitive in perceptual quality.

\noindent\textbf{RealSR.} As shown in Table~\ref{tab:main_realsr}, our method performs more favorably on real-world data. It achieves the best MANIQA and CLIPIQA, the second-best PSNR and SSIM, and a highly competitive MUSIQ. Compared with recent one-step baselines, our method provides a stronger overall balance between fidelity and perceptual quality on authentic degradations.


\textbf{Efficiency.} Table~\ref{tab:efficiency_drealsr_horizontal} compares the efficiency of different SR methods. 
Our one-step method further reduces end-to-end latency and MACs to 0.019\,s and 407.95\,G, respectively. 
Compared with the strongest recent one-step baseline AdcSR, our method is faster (0.019\,s vs.\ 0.03\,s), lighter in computation (407.95\,G vs.\ 496\,G MACs), and also uses fewer parameters (344\,M vs.\ 456\,M).

\begin{figure}[t]
    \vspace{-2mm}
	\centering
	\includegraphics[width=\linewidth]{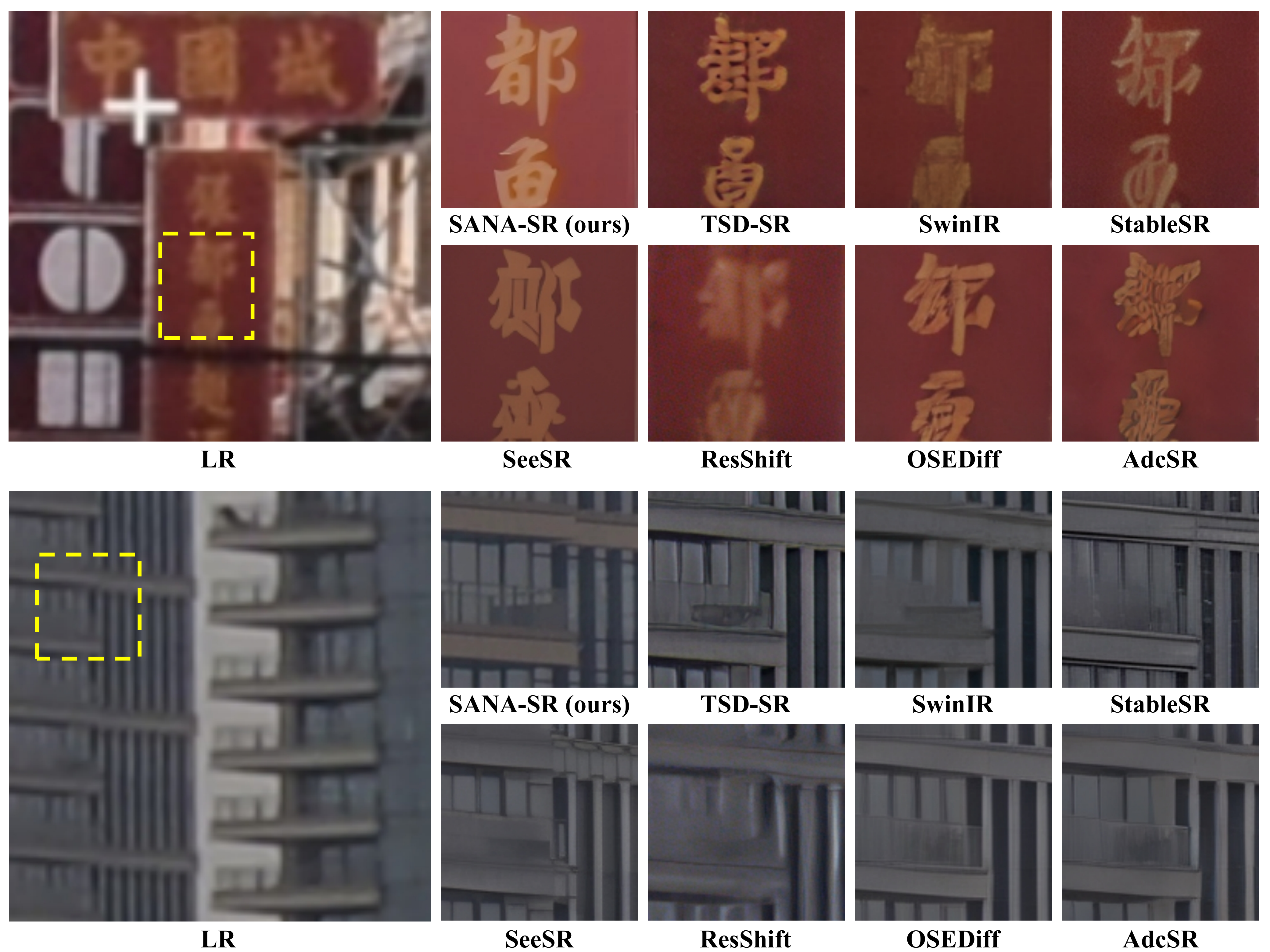}
    \vspace{-7mm}
	\caption{Qualitative comparison on challenging examples from DRealSR.}
	\label{fig:qualitative_results}
\end{figure}

\noindent\textbf{Qualitative comparison.} Fig.~\ref{fig:qualitative_results} shows that the efficiency gain of SANA-SR does not come at the expense of visual quality. 
\cao{(First example) our method recovers clearer and more recognizable character structures, whereas several baselines fail to recover readable high-frequency details. 
(Second example) SANA-SR restores sharper window boundaries and more coherent local structures on the building facade, while other methods tend to generate blurrier edges or less stable geometric details. }

\begin{table*}[t]
\vspace{-1mm}
\centering
\small
\setlength{\tabcolsep}{4.5pt}
\renewcommand{\arraystretch}{1.10}
\caption{Efficiency comparison of different kinds of methods.
}
\label{tab:efficiency_drealsr_horizontal}
\resizebox{\textwidth}{!}{%
\begin{tabular}{lcccccccccc}
\toprule
& \multicolumn{5}{c}{Prior-based multi-step / flexible-step} & \multicolumn{5}{c}{One-step methods} \\
\cmidrule(lr){2-6}\cmidrule(lr){7-11}
Metric & StableSR & DiffBIR & SeeSR & PASD & ResShift & SinSR & OSEDiff & S3Diff & AdcSR & \textbf{Ours} \\
\midrule
\#Steps$\downarrow$ 
& 200 & 50 & 50 & 20 & 15 & \textbf{1} & \textbf{1} & \textbf{1} & \textbf{1} & \textbf{1} \\

Time (s)$\downarrow$ 
& 11.50 & 2.72 & 4.30 & 2.80 & 0.71 & 0.13 & 0.11 & 0.28 & 0.03 & \textbf{0.019} \\

MACs (G)$\downarrow$ 
& 79,940 & 24,234 & 65,857 & 29,125 & 5,491 & 2,649 & 2,265 & 2,627 & 496 & \textbf{407.95} \\

\#Param. (M)$\downarrow$ 
& 1,410 & 1,717 & 2,524 & 1,900 & \textbf{119} & \textbf{119} & 1,775 & 1,327 & 456 & 344 \\
\bottomrule
\end{tabular}%
}
\vspace{-3mm}
\end{table*}

\subsection{Ablation Studies}
\label{sec:ablation}
\begin{figure}[t]
    \centering
    \vspace{-1mm}
    \includegraphics[width=\linewidth]{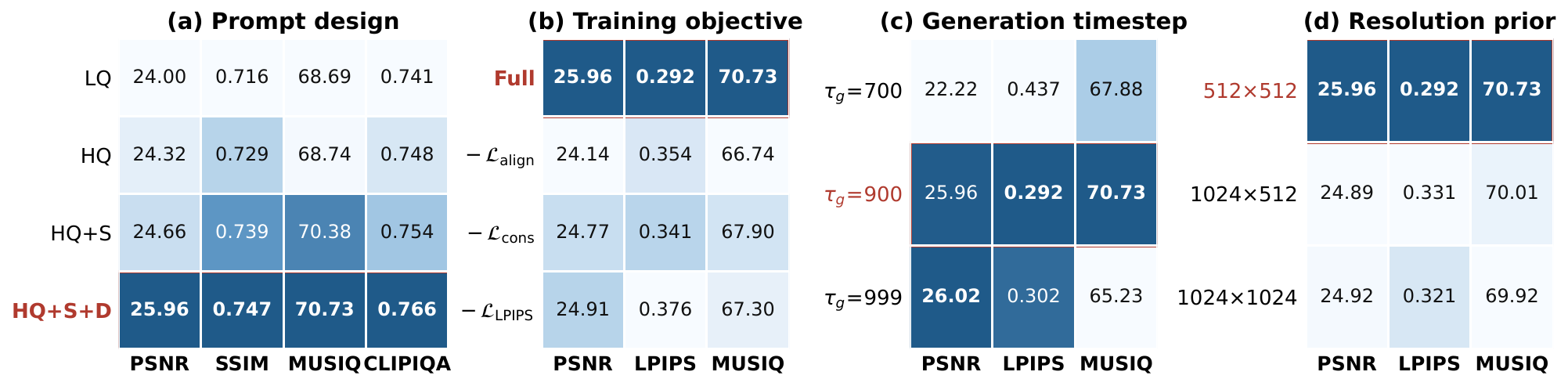}
    \includegraphics[width=\linewidth]{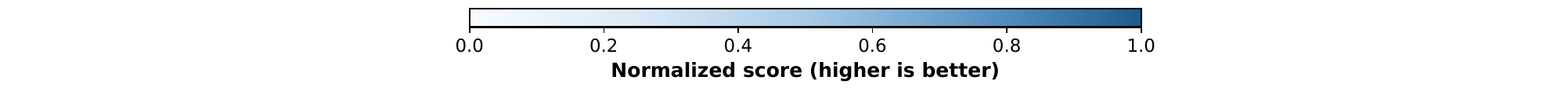}
    \vspace{-7mm}
    \caption{Ablation visualization of SANA-SR.
    Cell color encodes the per-panel,
    per-metric normalized score, with LPIPS reversed so that darker always
    denotes better performance, and our default configuration is outlined in \best{red}.
    \textbf{(a)} Prompt design, \textbf{S}: the quality suffix tag, \textbf{D}: DAPE prompts. 
    \textbf{(b)} Training objective: ``Full'' is our full objective and
    $-\mathcal{L}_{x}$ removes the corresponding loss term.
    \textbf{(c)} One-step generation timestep $\tau_g$. 
    \textbf{(d)} Resolution prior, denoted as ``backbone $\times$ training crop''. 
    }
    \label{tab:ablation_core}
    \vspace{-3mm}
\end{figure}

\vspace{-2mm}
\textbf{Prompt design.}
As shown in Fig~\ref{tab:ablation_core}, prompt design has a consistent impact on both fidelity and perceptual quality. Moving from an LQ prompt to an HQ prompt already improves PSNR/SSIM, and adding the quality suffix further yields a clear gain in MUSIQ. DAPE gives the best overall result, improving all reported metrics and producing the largest gain on CLIPIQA, which suggests that better prompt construction mainly benefits perceptual realism and semantic consistency.


\textbf{Training objective.}
Fig.~\ref{tab:ablation_core} shows that frozen-prior alignment is the most critical component in training. Removing $\mathcal{L}_{\mathrm{align}}$ causes the largest drop in PSNR, LPIPS, and MUSIQ, showing that image-space supervision alone is insufficient for stable one-step restoration. Removing $\mathcal{L}_{\mathrm{cons}}$ also hurts performance, but less severely, indicating that adapter consistency provides complementary regularization. By contrast, removing LPIPS leaves PSNR nearly unchanged but significantly degrades perceptual quality, confirming that pixel-wise supervision alone is not enough for Real-ISR.

\textbf{Generation timestep and resolution prior.}
Fig.~\ref{tab:ablation_core} also shows that the generation timestep controls a clear fidelity--perception trade-off. A smaller timestep under-restores the image, while an extremely late timestep slightly improves PSNR/LPIPS but causes a large MUSIQ drop. We therefore choose $\tau_g=900$ as the best balance. The resolution-prior ablation in Fig.~\ref{tab:ablation_core} further shows that replacing SANA-512 with SANA-1024 does not bring consistent gains, suggesting that performance mainly comes from effective adaptation rather than simply scaling the prior.

\begin{wraptable}[8]{r}{0.55\textwidth}
    \centering
    \vspace{-5mm}
    \small
    \setlength{\tabcolsep}{4pt}
    \caption{Comparisons of different pruning strategies.}
    \label{tab:ablation_prune}
    \resizebox{1\linewidth}{!}{
    \begin{tabular}{lccccc} 
    \toprule
    Strategy & Params & Time(s)$\downarrow$ & MACs$\downarrow$ & PSNR$\uparrow$ & MUSIQ$\uparrow$ \\
    \midrule
    Full model                  & full & 0.055 & 749.48 & 25.9638 & 70.7325 \\
    Tail pruning              & 0.350B & 0.028 & 474.31 & 24.7866 & 64.1776 \\
    Random pruning            & 0.323B & 0.017 & 383.05 & 22.2923 & 52.4731 \\
    Unstructured pruning & $\sim$0.35B & 0.034 & 463.31 & 24.7135 & 58.8295 \\
    Ours                      & 0.344B & 0.019 & 407.95 & \textbf{25.4923} & \textbf{70.6861} \\
    \bottomrule
    \end{tabular}
    }
\end{wraptable}
\textbf{Prompt-aware structured pruning.}
As reported in Table~\ref{tab:ablation_prune}, our pruning strategy achieves the best quality--efficiency trade-off under the 0.35B budget. It reduces latency and MACs substantially while preserving most of the restoration quality. In contrast, other pruning strategies lead to much larger degradation, especially in perceptual quality. This indicates that effective compression for one-step SR should preserve prompt-conditioned restoration behavior instead of relying on generic pruning heuristics.

\vspace{-3mm}
\section{Conclusion}
\label{sec:conclusion}
\vspace{-4mm}
\cao{We presented SANA-SR, an efficient one-step framework for Real-ISR. Motivated by the observation that existing one-step diffusion-based or flow matching based methods still suffer from dense latent representations and costly HR token interactions, we revisited the task from the perspectives of compact latent representation and efficient token modeling. To this end, SANA-SR combines a degradation-stable deep compression autoencoder, a degradation-aware prompted one-step LinearDiT adapted from a pretrained SANA prior, frozen-prior alignment and adapter consistency for stable one-step adaptation, and prompt-aware structured pruning for efficient deployment. As a result, SANA-SR achieves $32{\times}$ compression ratio and linear complexity.
Extensive experiments on both synthetic and real-image benchmarks show that SANA-SR achieves a favorable trade-off between restoration quality and efficiency, delivering competitive or superior quantitative results while running in 0.019s with 407.95G MACs and 344M parameters. }


\clearpage
\begin{ack}
This work was partially supported by the National Natural Science Fund for Excellent Young Scientists Fund Program (Overseas).
\end{ack}

\bibliography{ref}  

@article{wang2024stablesr,
  title={Exploiting diffusion prior for real-world image super-resolution},
  author={Wang, Jianyi and Yue, Zongsheng and Zhou, Shangchen and Chan, Kelvin CK and Loy, Chen Change},
  journal={International Journal of Computer Vision},
  volume={132},
  number={12},
  pages={5929--5949},
  year={2024},
  publisher={Springer}
}

@inproceedings{wu2024seesr,
  title={Seesr: Towards semantics-aware real-world image super-resolution},
  author={Wu, Rongyuan and Yang, Tao and Sun, Lingchen and Zhang, Zhengqiang and Li, Shuai and Zhang, Lei},
  booktitle={Proceedings of the IEEE/CVF Conference on Computer Vision and Pattern Recognition},
  pages={25456--25467},
  year={2024}
}

@inproceedings{yu2024supir,
  title={Scaling up to excellence: Practicing model scaling for photo-realistic image restoration in the wild},
  author={Yu, Fanghua and Gu, Jinjin and Li, Zheyuan and Hu, Jinfan and Kong, Xiangtao and Wang, Xintao and He, Jingwen and Qiao, Yu and Dong, Chao},
  booktitle={Proceedings of the IEEE/CVF Conference on Computer Vision and Pattern Recognition},
  pages={25669--25680},
  year={2024}
}

@inproceedings{lin2024diffbir,
  title={Diffbir: Toward blind image restoration with generative diffusion prior},
  author={Lin, Xinqi and He, Jingwen and Chen, Ziyan and Lyu, Zhaoyang and Dai, Bo and Yu, Fanghua and Qiao, Yu and Ouyang, Wanli and Dong, Chao},
  booktitle={European Conference on Computer Vision},
  pages={430--448},
  year={2024},
  organization={Springer}
}

@article{ai2024dreamclear,
  title={Dreamclear: High-capacity real-world image restoration with privacy-safe dataset curation},
  author={Ai, Yuang and Zhou, Xiaoqiang and Huang, Huaibo and Han, Xiaotian and Chen, Zhengyu and You, Quanzeng and Yang, Hongxia},
  journal={Advances in Neural Information Processing Systems},
  volume={37},
  pages={55443--55469},
  year={2024}
}

@inproceedings{duan2025dit4sr,
  title={Dit4sr: Taming diffusion transformer for real-world image super-resolution},
  author={Duan, Zheng-Peng and Zhang, Jiawei and Jin, Xin and Zhang, Ziheng and Xiong, Zheng and Zou, Dongqing and Ren, Jimmy S and Guo, Chunle and Li, Chongyi},
  booktitle={Proceedings of the IEEE/CVF International Conference on Computer Vision},
  pages={18948--18958},
  year={2025}
}

@inproceedings{wang2024sinsr,
  title={Sinsr: diffusion-based image super-resolution in a single step},
  author={Wang, Yufei and Yang, Wenhan and Chen, Xinyuan and Wang, Yaohui and Guo, Lanqing and Chau, Lap-Pui and Liu, Ziwei and Qiao, Yu and Kot, Alex C and Wen, Bihan},
  booktitle={Proceedings of the IEEE/CVF Conference on Computer Vision and Pattern Recognition},
  pages={25796--25805},
  year={2024}
}

@article{wu2024osediff,
  title={One-step effective diffusion network for real-world image super-resolution},
  author={Wu, Rongyuan and Sun, Lingchen and Ma, Zhiyuan and Zhang, Lei},
  journal={Advances in Neural Information Processing Systems},
  volume={37},
  pages={92529--92553},
  year={2024}
}

@inproceedings{dong2025tsdsr,
  title={Tsd-sr: One-step diffusion with target score distillation for real-world image super-resolution},
  author={Dong, Linwei and Fan, Qingnan and Guo, Yihong and Wang, Zhonghao and Zhang, Qi and Chen, Jinwei and Luo, Yawei and Zou, Changqing},
  booktitle={Proceedings of the IEEE/CVF Conference on Computer Vision and Pattern Recognition},
  pages={23174--23184},
  year={2025}
}

@inproceedings{yue2025invsr,
  title={Arbitrary-steps image super-resolution via diffusion inversion},
  author={Yue, Zongsheng and Liao, Kang and Loy, Chen Change},
  booktitle={Proceedings of the IEEE/CVF Conference on Computer Vision and Pattern Recognition},
  pages={23153--23163},
  year={2025}
}

@inproceedings{sun2025pisasr,
  title={Pixel-level and semantic-level adjustable super-resolution: A dual-lora approach},
  author={Sun, Lingchen and Wu, Rongyuan and Ma, Zhiyuan and Liu, Shuaizheng and Yi, Qiaosi and Zhang, Lei},
  booktitle={Proceedings of the IEEE/CVF Conference on Computer Vision and Pattern Recognition},
  pages={2333--2343},
  year={2025}
}

@inproceedings{chen2025adcsr,
  title={Adversarial diffusion compression for real-world image super-resolution},
  author={Chen, Bin and Li, Gehui and Wu, Rongyuan and Zhang, Xindong and Chen, Jie and Zhang, Jian and Zhang, Lei},
  booktitle={Proceedings of the IEEE/CVF Conference on Computer Vision and Pattern Recognition},
  pages={28208--28220},
  year={2025}
}

@inproceedings{li2025d3sr,
  title={Unleashing the Power of One-Step Diffusion based Image Super-Resolution via a Large-Scale Diffusion Discriminator},
  author={Li, Jianze and Cao, Jiezhang and Zou, Zichen and Su, Xiongfei and Yuan, Xin and Zhang, Yulun and Guo, Yong and Yang, Xiaokang},
  booktitle={Advances in Neural Information Processing Systems},
  year = {2025}
}

@inproceedings{xie2025sana,
  title={Sana: Efficient high-resolution image synthesis with linear diffusion transformers},
  author={Xie, Enze and Chen, Junsong and Chen, Junyu and Cai, Han and Tang, Haotian and Lin, Yujun and Zhang, Zhekai and Li, Muyang and Zhu, Ligeng and Lu, Yao and others},
  booktitle={International Conference on Learning Representations},
  year={2025}
}

@inproceedings{li2026linearsr,
  title     = {{LinearSR}: Unlocking Linear Attention for Stable and Efficient Image Super-Resolution},
  author    = {Li, Xiaohui and Zhuang, Shaobin and Cao, Shuo and Yang, Yang and Pu, Yuandong and Qin, Qi and Luo, Siqi and Fu, Bin and Liu, Yihao},
  booktitle = {International Conference on Learning Representations},
  year      = {2026}
}

@article{zhang2026qdit4sr,
  title={Q-DiT4SR: Exploration of Detail-Preserving Diffusion Transformer Quantization for Real-World Image Super-Resolution},
  author={Zhang, Xun and Yang, Kaicheng and Lu, Hongliang and Qin, Haotong and Guo, Yong and Zhang, Yulun},
  journal={arXiv preprint arXiv:2602.01273},
  year={2026}
}

@article{cui2024dossr,
  title={Taming Diffusion Prior for Image Super-Resolution with Domain Shift SDEs},
  author={Cui, Qinpeng and Liu, Yixuan and Zhang, Xinyi and Bao, Qiqi and Liao, Qingmin and Wang, Li and Lu, Tian and Wang, Zhongdao and Barsoum, Emad and others},
  journal={Advances in Neural Information Processing Systems},
  volume={37},
  pages={42765--42797},
  year={2024}
}

@inproceedings{ignatov2025quantized,
  title={Quantized image super-resolution on mobile npus, mobile ai 2025 challenge: Report},
  author={Ignatov, Andrey and Perevozchikov, Georgy and Timofte, Radu and Zhang, Zhiyu and Gao, Tianxiao and Yang, Yukun and Zhu, Shiai and Wang, Shihao and Yoon, Kihwan and Gankhuyag, Ganzorig and others},
  booktitle={Proceedings of the IEEE/CVF Conference on Computer Vision and Pattern Recognition Workshops},
  pages={1908--1921},
  year={2025}
}

@article{wang2020deep,
  title={Deep learning for image super-resolution: A survey},
  author={Wang, Zhihao and Chen, Jian and Hoi, Steven CH},
  journal={IEEE Transactions on Pattern Analysis and Machine Intelligence},
  volume={43},
  number={10},
  pages={3365--3387},
  year={2020},
  publisher={IEEE}
}

@inproceedings{lugmayr2020ntire,
  title={Ntire 2020 challenge on real-world image super-resolution: Methods and results},
  author={Lugmayr, Andreas and Danelljan, Martin and Timofte, Radu},
  booktitle={Proceedings of the IEEE/CVF Conference on Computer Vision and Pattern Recognition Workshops},
  pages={494--495},
  year={2020}
}

@inproceedings{xiao2026reversible,
  title={Reversible primitive--composition alignment for continual vision--language learning},
  author={Xiao, Canran and Xu, Tianxiang and Ma, Siyuan and Jiang, Yiyang and Gao, Haoyu and Wu, Yuhan},
  booktitle={International Conference on Learning Representations},
  year={2026}
}

@inproceedings{yang2024pixel,
  title={Pixel-aware stable diffusion for realistic image super-resolution and personalized stylization},
  author={Yang, Tao and Wu, Rongyuan and Ren, Peiran and Xie, Xuansong and Zhang, Lei},
  booktitle={European Conference on Computer Vision},
  pages={74--91},
  year={2024},
  organization={Springer}
}

@article{yue2023resshift,
  title={Resshift: Efficient diffusion model for image super-resolution by residual shifting},
  author={Yue, Zongsheng and Wang, Jianyi and Loy, Chen Change},
  journal={Advances in Neural Information Processing Systems},
  volume={36},
  pages={13294--13307},
  year={2023}
}

@article{zhang2024degradation,
  title={Degradation-guided one-step image super-resolution with diffusion priors},
  author={Zhang, Aiping and Yue, Zongsheng and Pei, Renjing and Ren, Wenqi and Cao, Xiaochun},
  journal={arXiv preprint arXiv:2409.17058},
  year={2024}
}

@article{tai2026addsr,
  title={Addsr: Accelerating diffusion-based blind super-resolution with adversarial diffusion distillation},
  author={Tai, Ying and Xie, Rui and Zhao, Chen and Zhang, Kai and Zhang, Zhenyu and Zhou, Jun and Yang, Jian},
  journal={Pattern Recognition},
  pages={113012},
  year={2026},
  publisher={Elsevier}
}

@inproceedings{li2025one,
  title={One Diffusion Step to Real-World Super-Resolution via Flow Trajectory Distillation},
  author={Li, Jianze and Cao, Jiezhang and Guo, Yong and Li, Wenbo and Zhang, Yulun},
  booktitle={International Conference on Machine Learning},
  pages={34044--34053},
  year={2025},
  organization={PMLR}
}

@inproceedings{agustsson2017ntire,
  title={Ntire 2017 challenge on single image super-resolution: Dataset and study},
  author={Agustsson, Eirikur and Timofte, Radu},
  booktitle={Proceedings of the IEEE/CVF Conference on Computer Vision and Pattern Recognition Workshops},
  pages={126--135},
  year={2017}
}

@inproceedings{cai2019toward,
  title={Toward real-world single image super-resolution: A new benchmark and a new model},
  author={Cai, Jianrui and Zeng, Hui and Yong, Hongwei and Cao, Zisheng and Zhang, Lei},
  booktitle={Proceedings of the IEEE/CVF International Conference on Computer Vision},
  pages={3086--3095},
  year={2019}
}

@inproceedings{wei2020component,
  title={Component divide-and-conquer for real-world image super-resolution},
  author={Wei, Pengxu and Xie, Ziwei and Lu, Hannan and Zhan, Zongyuan and Ye, Qixiang and Zuo, Wangmeng and Lin, Liang},
  booktitle={European Conference on Computer Vision},
  pages={101--117},
  year={2020},
  organization={Springer}
}

@inproceedings{wang2021real,
  title={Real-esrgan: Training real-world blind super-resolution with pure synthetic data},
  author={Wang, Xintao and Xie, Liangbin and Dong, Chao and Shan, Ying},
  booktitle={Proceedings of the IEEE/CVF International Conference on Computer Vision},
  pages={1905--1914},
  year={2021}
}

@inproceedings{dong2014learning,
  title={Learning a deep convolutional network for image super-resolution},
  author={Dong, Chao and Loy, Chen Change and He, Kaiming and Tang, Xiaoou},
  booktitle={European Conference on Computer Vision},
  pages={184--199},
  year={2014},
  organization={Springer}
}

@inproceedings{lim2017enhanced,
  title={Enhanced deep residual networks for single image super-resolution},
  author={Lim, Bee and Son, Sanghyun and Kim, Heewon and Nah, Seungjun and Mu Lee, Kyoung},
  booktitle={Proceedings of the IEEE/CVF Conference on Computer Vision and Pattern Recognition Workshops},
  pages={136--144},
  year={2017}
}

@inproceedings{zhang2018image,
  title={Image super-resolution using very deep residual channel attention networks},
  author={Zhang, Yulun and Li, Kunpeng and Li, Kai and Wang, Lichen and Zhong, Bineng and Fu, Yun},
  booktitle={European Conference on Computer Vision},
  pages={286--301},
  year={2018}
}

@inproceedings{wang2018esrgan,
  title={Esrgan: Enhanced super-resolution generative adversarial networks},
  author={Wang, Xintao and Yu, Ke and Wu, Shixiang and Gu, Jinjin and Liu, Yihao and Dong, Chao and Qiao, Yu and Change Loy, Chen},
  booktitle={European Conference on Computer Vision Workshops},
  year={2018}
}

@inproceedings{liang2021swinir,
  title={Swinir: Image restoration using swin transformer},
  author={Liang, Jingyun and Cao, Jiezhang and Sun, Guolei and Zhang, Kai and Van Gool, Luc and Timofte, Radu},
  booktitle={Proceedings of the IEEE/CVF International Conference on Computer Vision},
  pages={1833--1844},
  year={2021}
}

@inproceedings{chen2023activating,
  title={Activating more pixels in image super-resolution transformer},
  author={Chen, Xiangyu and Wang, Xintao and Zhou, Jiantao and Qiao, Yu and Dong, Chao},
  booktitle={Proceedings of the IEEE/CVF Conference on Computer Vision and Pattern Recognition},
  pages={22367--22377},
  year={2023}
}

@inproceedings{zhang2021designing,
  title={Designing a practical degradation model for deep blind image super-resolution},
  author={Zhang, Kai and Liang, Jingyun and Van Gool, Luc and Timofte, Radu},
  booktitle={Proceedings of the IEEE/CVF International Conference on Computer Vision},
  pages={4791--4800},
  year={2021}
}

@article{lecun1989optimal,
  title={Optimal brain damage},
  author={LeCun, Yann and Denker, John and Solla, Sara},
  journal={Advances in Neural Information Processing Systems},
  volume={2},
  year={1989}
}

@article{han2015learning,
  title={Learning both weights and connections for efficient neural network},
  author={Han, Song and Pool, Jeff and Tran, John and Dally, William},
  journal={Advances in Neural Information Processing Systems},
  volume={28},
  year={2015}
}

@inproceedings{liu2017learning,
  title={Learning efficient convolutional networks through network slimming},
  author={Liu, Zhuang and Li, Jianguo and Shen, Zhiqiang and Huang, Gao and Yan, Shoumeng and Zhang, Changshui},
  booktitle={Proceedings of the IEEE/CVF International Conference on Computer Vision},
  pages={2736--2744},
  year={2017}
}

@inproceedings{frankle2018lottery,
  title={The lottery ticket hypothesis: Finding sparse, trainable neural networks},
  author={Frankle, Jonathan and Carbin, Michael},
  booktitle={International Conference on Learning Representations},
  year={2019}
}

@article{hoefler2021sparsity,
  title={Sparsity in deep learning: Pruning and growth for efficient inference and training in neural networks},
  author={Hoefler, Torsten and Alistarh, Dan and Ben-Nun, Tal and Dryden, Nikoli and Peste, Alexandra},
  journal={Journal of Machine Learning Research},
  volume={22},
  number={241},
  pages={1--124},
  year={2021}
}

@inproceedings{fang2023depgraph,
  title={Depgraph: Towards any structural pruning},
  author={Fang, Gongfan and Ma, Xinyin and Song, Mingli and Mi, Michael Bi and Wang, Xinchao},
  booktitle={Proceedings of the IEEE/CVF Conference on Computer Vision and Pattern Recognition},
  pages={16091--16101},
  year={2023}
}

@article{ma2023llm,
  title={Llm-pruner: On the structural pruning of large language models},
  author={Ma, Xinyin and Fang, Gongfan and Wang, Xinchao},
  journal={Advances in Neural Information Processing Systems},
  volume={36},
  pages={21702--21720},
  year={2023}
}

@inproceedings{fang2025tinyfusion,
  title={Tinyfusion: Diffusion transformers learned shallow},
  author={Fang, Gongfan and Li, Kunjun and Ma, Xinyin and Wang, Xinchao},
  booktitle={Proceedings of the IEEE/CVF Conference on Computer Vision and Pattern Recognition},
  pages={18144--18154},
  year={2025}
}

@inproceedings{rombach2022high,
  title={High-resolution image synthesis with latent diffusion models},
  author={Rombach, Robin and Blattmann, Andreas and Lorenz, Dominik and Esser, Patrick and Ommer, Bj{\"o}rn},
  booktitle={Proceedings of the IEEE/CVF Conference on Computer Vision and Pattern Recognition},
  pages={10684--10695},
  year={2022}
}

@inproceedings{peebles2023scalable,
  title={Scalable diffusion models with transformers},
  author={Peebles, William and Xie, Saining},
  booktitle={Proceedings of the IEEE/CVF Conference on Computer Vision and Pattern Recognition},
  pages={4195--4205},
  year={2023}
}

@inproceedings{katharopoulos2020transformers,
  title={Transformers are rnns: Fast autoregressive transformers with linear attention},
  author={Katharopoulos, Angelos and Vyas, Apoorv and Pappas, Nikolaos and Fleuret, Fran{\c{c}}ois},
  booktitle={International Conference on Machine Learning},
  pages={5156--5165},
  year={2020},
  organization={PMLR}
}

@inproceedings{hu2022lora,
  title={{LoRA}: Low-Rank Adaptation of Large Language Models},
  author={Hu, Edward J and Shen, Yelong and Wallis, Phillip and Allen-Zhu, Zeyuan and Li, Yuanzhi and Wang, Shean and Wang, Lu and Chen, Weizhu},
  booktitle={International Conference on Learning Representations},
  year={2022}
}

@inproceedings{li2023lsdir,
  title={Lsdir: A large scale dataset for image restoration},
  author={Li, Yawei and Zhang, Kai and Liang, Jingyun and Cao, Jiezhang and Liu, Ce and Gong, Rui and Zhang, Yulun and Tang, Hao and Liu, Yun and Demandolx, Denis and others},
  booktitle={Proceedings of the IEEE/CVF Conference on Computer Vision and Pattern Recognition},
  pages={1775--1787},
  year={2023}
}

@inproceedings{karras2019style,
  title={A style-based generator architecture for generative adversarial networks},
  author={Karras, Tero and Laine, Samuli and Aila, Timo},
  booktitle={Proceedings of the IEEE/CVF Conference on Computer Vision and Pattern Recognition},
  pages={4401--4410},
  year={2019}
}

@misc{labs2025flux1kontextflowmatching,
      title={FLUX.1 Kontext: Flow Matching for In-Context Image Generation and Editing in Latent Space},
      author={Black Forest Labs and Stephen Batifol and Andreas Blattmann and Frederic Boesel and Saksham Consul and Cyril Diagne and Tim Dockhorn and Jack English and Zion English and Patrick Esser and Sumith Kulal and Kyle Lacey and Yam Levi and Cheng Li and Dominik Lorenz and Jonas Müller and Dustin Podell and Robin Rombach and Harry Saini and Axel Sauer and Luke Smith},
      year={2025},
      eprint={2506.15742},
      archivePrefix={arXiv},
      primaryClass={cs.GR},
      url={https://arxiv.org/abs/2506.15742},
}

@misc{flux2024,
    author={Black Forest Labs},
    title={FLUX},
    year={2024},
    howpublished={\url{https://github.com/black-forest-labs/flux}},
}


\appendix
\newpage
\section*{Appendix}

\section{Additional Technical Details}
\label{app:technical}

\subsection{Prompt Construction and Protocol}
\label{app:prompt}

For every experimental run, the prompt source is fixed once and used consistently across training, validation, and pruning calibration. Let $x_L$ denote that prompt image. The targeted prompt extractor $\Pi(\cdot)$ produces an image-dependent tag string, which is concatenated with a fixed quality template $p_{\mathrm{tpl}}$:
\begin{equation}
p = \Pi(x_L)\oplus p_{\mathrm{tpl}},
\qquad
(c,m)=\mathcal{T}(p),
\label{eq:app_prompt}
\end{equation}
where $\oplus$ denotes string concatenation, $\mathcal{T}$ is the frozen SANA tokenizer-text encoder, $c\in\mathbb{R}^{L\times d_t}$ is the token embedding sequence, and $m\in\{0,1\}^{L}$ is the text-attention mask. All image restoration, validation, and pruning decisions are conditioned on the same prompt construction rule in Eq.~\eqref{eq:app_prompt}.

\subsection{LoRA Adaptation of the LinearDiT Backbone}
\label{app:lora}

Let $f_0$ denote the pretrained SANA LinearDiT backbone. We inject LoRA updates into its attention projections and keep the original pretrained weights frozen. For any target projection matrix $W_0\in\mathbb{R}^{d_{\mathrm{out}}\times d_{\mathrm{in}}}$, the adapted weight is
\begin{equation}
W
=
W_0 + \Delta W,
\qquad
\Delta W
=
\frac{\alpha_{\mathrm{LoRA}}}{r}BA,
\label{eq:app_lora}
\end{equation}
where $A\in\mathbb{R}^{r\times d_{\mathrm{in}}}$ and $B\in\mathbb{R}^{d_{\mathrm{out}}\times r}$ are trainable low-rank factors, $r$ is the LoRA rank, and $\alpha_{\mathrm{LoRA}}$ is the LoRA scaling coefficient. We denote the resulting adapted backbone by $f_\theta$.

\subsection{Native LinearDiT Backbone}
\label{app:lineardit}

SANA is built on a LinearDiT architecture. For completeness, a generic linear-attention block can be written as
\begin{equation}
\operatorname{LA}(Q,K,V)
=
\frac{\phi(Q)\big(\phi(K)^\top V\big)}
{\phi(Q)\big(\phi(K)^\top \mathbf{1}\big)+\varepsilon_{\mathrm{att}}},
\label{eq:app_la}
\end{equation}
where $Q,K\in\mathbb{R}^{N\times d}$, $V\in\mathbb{R}^{N\times d_v}$, $\phi(\cdot)$ is a non-negative feature map, $\mathbf{1}\in\mathbb{R}^{N}$ is the all-one vector, and $\varepsilon_{\mathrm{att}}>0$ is a stabilizer. Equation~\eqref{eq:app_la} is included only to make the backbone assumption explicit; SANA-SR itself does not modify the internal LinearDiT operator, and adapts it solely through LoRA.

\paragraph{Asymptotic complexity of representative Real-ISR families.}
Combined with the $32\times$ DC-AE in \S\ref{sec:method:dcae}, the linear-attention LinearDiT operator (Eq.~\eqref{eq:app_la}) gives SANA-SR an asymptotic per-image cost of $\mathcal{O}(N d^2)$ with $N\!\approx\!256$ at a $512\times512$ working resolution. By contrast, prior-driven multi-step restorers such as StableSR, DiffBIR, and SeeSR scale as $\mathcal{O}(K\, N^2 d)$ with $K\!=\!20$--$200$ sampling steps, SD-based one-step methods such as OSEDiff, AdcSR, and TSD-SR retain the $\mathcal{O}(N^2 d)$ per-step cost at $N\!\approx\!4{,}096$, and recent multi-step LinearDiT restorers such as LinearSR scale as $\mathcal{O}(K\, N d^2)$. SANA-SR therefore reduces $N$ by roughly $16\times$ through deep latent compression, replaces quadratic self-attention with linear attention, and collapses the sampling depth $K$ to a single update; these three multiplicative effects account for the large reduction in MACs and end-to-end latency reported in Table~\ref{tab:efficiency_drealsr_horizontal} and visualized in Fig.~\ref{fig:framework}.

\subsection{Latent Encoding, Decoding, and Scheduler Coefficients}
\label{app:latent}

All latent operations follow the native SANA VAE convention. The frozen encoder and decoder are denoted by $E$ and $D$, respectively:
\begin{equation}
z_L = E(x_L), \qquad z_H = E(x_H), \qquad \hat x = D(\hat z).
\label{eq:app_vae}
\end{equation}
The one-step generation timestep $\tau_g$ is fixed during training and inference, while the alignment timestep $t$ is sampled uniformly from a prescribed training range:
\begin{equation}
t \sim \mathcal{U}\{t_{\min}, t_{\min}+1,\dots,t_{\max}\},
\qquad
\epsilon \sim \mathcal{N}(0,I).
\label{eq:app_time}
\end{equation}
The perturbed latents use the scheduler coefficients $(\alpha_t,\sigma_t)$ from the SANA scheduler:
\begin{equation}
\tilde z_{\hat z} = \alpha_t \hat z + \sigma_t \epsilon,
\qquad
\tilde z_H = \alpha_t z_H + \sigma_t \epsilon.
\label{eq:app_noisy}
\end{equation}
In the current implementation, the scheduler explicitly materializes $\sigma_t$, and the perturbation is instantiated with $\alpha_t=1-\sigma_t$.

\subsection{Exact Frozen-Prior Alignment Computation}
\label{app:align}

The frozen-prior alignment branch evaluates the same frozen backbone on the perturbed restored and reference latents:
\begin{equation}
q_{\hat z}=f_0(\tilde z_{\hat z},t,c,m),
\qquad
q_H=f_0(\tilde z_H,t,c,m).
\label{eq:app_probe}
\end{equation}
For a response tensor $q\in\mathbb{R}^{B\times C\times h\times w}$, we define
\begin{equation}
\mu_{b,c}(q)
=
\frac{1}{hw}\sum_{u=1}^{h}\sum_{v=1}^{w} q_{b,c,u,v},
\qquad
s_{b,c}(q)
=
\frac{1}{hw}\sum_{u=1}^{h}\sum_{v=1}^{w}
\big(q_{b,c,u,v}-\mu_{b,c}(q)\big)^2.
\label{eq:app_stats}
\end{equation}
The alignment loss in the main text is then the channel-wise KL divergence between the Gaussian summaries defined by Eq.~\eqref{eq:app_stats}. The consistency branch evaluates
\[
q_{\hat z}^{\mathrm{adapt}} = f_\theta(\tilde z_{\hat z},t,c,m),
\qquad
q_{\hat z}^{\mathrm{base}} = f_0(\tilde z_{\hat z},t,c,m),
\]
and penalizes their mean-squared deviation.

\subsection{Exact Training Objective}
\label{app:objective}

The full training objective can be written explicitly as
\begin{equation}
\mathcal{L}
=
\lambda_2\frac{1}{3HW}\|\hat x-x_H\|_2^2
+
\lambda_p\operatorname{LPIPS}(\hat x,x_H)
+
\lambda_a\mathcal{L}_{\mathrm{align}}
+
\lambda_c\mathcal{L}_{\mathrm{cons}}.
\label{eq:app_total}
\end{equation}
Only the LoRA parameters in $f_\theta$ are updated. The frozen VAE, frozen text encoder, and the base pretrained LinearDiT weights in $f_0$ are never optimized.

\subsection{Prompt-Aware Pruning Calibration}
\label{app:pruning}

Before pruning, we merge the learned LoRA adapters into the backbone and obtain a dense model $\bar f_\theta$. For calibration, we reuse the same prompt protocol as Eq.~\eqref{eq:app_prompt} and the same task loss as the main method, but we omit the adapter-consistency term:
\begin{equation}
\mathcal{L}_{\mathrm{cal}}
=
\omega(t)\Big(
\mathcal{L}_{\mathrm{rec}}+\lambda_a\mathcal{L}_{\mathrm{align}}
\Big).
\label{eq:app_cal}
\end{equation}
The timestep weight is
\begin{equation}
\omega(t)
=
\frac{\log(T+1)-\log(t+1)}{\log(T+1)},
\label{eq:app_weight}
\end{equation}
where $T$ is the total number of training timesteps. This weighting places larger emphasis on the timesteps that dominate the one-step restoration trajectory in the current implementation.

Let $K$ be the number of calibration iterations. The diagonal curvature proxy is
\begin{equation}
F_i
=
\frac{1}{K}
\sum_{k=1}^{K}
\omega(t_k)
\left(
\frac{\partial \mathcal{L}_{\mathrm{cal}}^{(k)}}{\partial w_i}
\right)^2.
\label{eq:app_fisher}
\end{equation}
For transformer block $\mathcal{B}_\ell$, the saliency is
\begin{equation}
S_\ell
=
\sum_{w_i \in \mathcal{B}_\ell}
\frac{w_i^2}{F_i+\varepsilon_p}.
\label{eq:app_saliency}
\end{equation}
We always retain the first and last transformer blocks. The remaining blocks are selected greedily by descending $S_\ell$ subject to the deployment budget.

\subsection{Training and Pruning Algorithms}
\label{app:algorithms}

\begin{algorithm}[H]
\caption{One-step SANA-SR training}
\label{alg:train_sanasr}
\begin{algorithmic}[1]
\Require paired batch $(x_L,x_H)$, frozen $E,D,\mathcal{T},f_0$, ...
\State construct prompt $p=\Pi(x_L)\oplus p_{\mathrm{tpl}}$ and encode $(c,m)=\mathcal{T}(p)$
\State encode latents $z_L=E(x_L)$ and $z_H=E(x_H)$
\State restore once: $\hat z = z_L - \sigma_{\tau_g} f_\theta(z_L,\tau_g,c,m)$
\State decode $\hat x=D(\hat z)$
\State sample $t \sim \mathcal{U}\{t_{\min},\dots,t_{\max}\}$ and $\epsilon\sim \mathcal{N}(0,I)$
\State form $\tilde z_{\hat z}=\alpha_t\hat z+\sigma_t\epsilon$ and $\tilde z_H=\alpha_t z_H+\sigma_t\epsilon$
\State compute $\mathcal{L}_{\mathrm{rec}}$, $\mathcal{L}_{\mathrm{align}}$, and $\mathcal{L}_{\mathrm{cons}}$
\State optimize LoRA parameters with $\mathcal{L}=\mathcal{L}_{\mathrm{rec}}+\lambda_a\mathcal{L}_{\mathrm{align}}+\lambda_c\mathcal{L}_{\mathrm{cons}}$
\end{algorithmic}
\end{algorithm}

\begin{algorithm}[H]
\caption{Prompt-aware structured pruning}
\label{alg:prune_sanasr}
\begin{algorithmic}[1]
\Require trained SANA-SR, calibration set, target budget $P_\star$
\State merge LoRA adapters into the LinearDiT backbone
\State split the LinearDiT into blocks $\{\mathcal{B}_\ell\}_{\ell=1}^{L}$
\For{$k=1$ to $K$}
    \State construct prompt-conditioned calibration tuple
    \State compute $\mathcal{L}_{\mathrm{cal}}^{(k)}=\omega(t_k)(\mathcal{L}_{\mathrm{rec}}^{(k)}+\lambda_a\mathcal{L}_{\mathrm{align}}^{(k)})$
    \State accumulate diagonal curvature proxy $F_i \leftarrow F_i + \omega(t_k)\left(\frac{\partial \mathcal{L}_{\mathrm{cal}}^{(k)}}{\partial w_i}\right)^2$
\EndFor
\State normalize $F_i$ by $K$
\State compute block saliency $S_\ell=\sum_{w_i\in\mathcal{B}_\ell}\frac{w_i^2}{F_i+\varepsilon_p}$
\State keep blocks by descending $S_\ell$ under budget $P_\star$, while always preserving the first and last blocks
\State remove the remaining blocks and save the pruned backbone
\end{algorithmic}
\end{algorithm}

\section{Extended Experimental Setup}
\label{app:exp_details}

\paragraph{Datasets and protocols.}
Recent diffusion-based Real-ISR works typically train on large high-quality corpora. Our SANA-SR is trained on a mixed HQ image pool of approximately $90$K images, combining DIV2K~\citep{agustsson2017ntire} ($800$), Flickr2K~\citep{lim2017enhanced} ($\sim$2.6K), LSDIR~\citep{li2023lsdir} ($\sim$85K), and FFHQ~\citep{karras2019style} (10K) training sets, deduplicated by SHA256. The released pipeline also supports any user-specified HQ directory, with optional RealSR~\citep{cai2019toward} HQ augmentation. When no precomputed LQ folder is provided, the code synthesizes degraded inputs online following the Real-ESRGAN degradation pipeline~\citep{wang2021real}. The main paper reports results on DIV2K-Val and RealSR, since they form the most common synthetic/real benchmark pair in recent work. DRealSR~\citep{wei2020component} is treated as an extended real-image benchmark and is evaluated with exactly the same inference and metric pipeline.

\paragraph{Evaluation metrics.}
We report two groups of metrics. First, PSNR and SSIM evaluate distortion fidelity; both are computed on the Y channel of YCbCr. Second, MANIQA, MUSIQ, CLIPIQA, and NIQE evaluate perceptual quality. The released evaluator supports both full-image evaluation and a $256\times256$ center-crop patch protocol. We use the full-image protocol by default and switch to the patch protocol only for fair comparison with methods whose official reports adopt crop-based evaluation.

\paragraph{Compared methods and fairness.}
Our comparison pool is built from methods that appear most frequently in recent works. Specifically, we include StableSR~\citep{wang2024stablesr}, DiffBIR~\citep{lin2024diffbir}, SeeSR~\citep{wu2024seesr}, SUPIR~\citep{yu2024supir}, DreamClear~\citep{ai2024dreamclear}, InvSR~\citep{yue2025invsr}, and LinearSR~\citep{li2026linearsr} as representative prior-driven multi-step or flexible-step methods, and SinSR~\citep{wang2024sinsr}, OSEDiff~\citep{wu2024osediff}, AdcSR~\citep{chen2025adcsr}, and TSD-SR~\citep{dong2025tsdsr} as fast or one-step competitors. We additionally regard ResShift~\citep{yue2023resshift} and PASD~\citep{yang2024pixel} as strong historical references. Since these methods differ substantially in architecture size, sampling trajectory length, and conditioning design, we do not over-interpret cross-category speed--quality trade-offs from a single score. Instead, we enforce fairness within each reported setting by using the same test split, the same upscaling factor, the same image-size handling, and the same evaluation script. 

\paragraph{Implementation details.}
All released training and evaluation scripts are 8*4090 scripts, and training is performed in FP16. Our codebase is implemented in PyTorch with Diffusers/Transformers for pretrained SANA components, PEFT for LoRA injection, and PyIQA for evaluation. We freeze the VAE, tokenizer/text encoder, and pretrained transformer weights outside LoRA. LoRA is inserted into \texttt{to\_q}, \texttt{to\_k}, \texttt{to\_v}, and \texttt{to\_out.0}, with rank 64, scale 64, and dropout 0. The default training setting uses batch size 4, $512\times512$ random crops, $4\times$ super-resolution, 12 dataloader workers, AdamW with learning rate $5\times10^{-5}$, weight decay $10^{-2}$, $(\beta_1,\beta_2)=(0.9,0.999)$, $\epsilon=10^{-8}$, gradient clipping at 1.0, 100K optimization steps, and EMA decay 0.999. The loss weights are $\lambda_2=1$, $\lambda_p=2$, $\lambda_a=1$, and $\lambda_c=1$.

\paragraph{Prompt protocol, checkpoint selection, and pruning.}
The released code supports both HQ-prompt and LQ-prompt protocols. Unless otherwise stated, we use the HQ-prompt protocol in the main paper. The HQ-prompt variant uses generation timestep $900$ and training-noise range $[70,650]$, whereas the LQ-prompt variant uses generation timestep $999$ and range $[20,980]$. Targeted prompts are extracted by the RAM/DAPE prompt module and appended with the suffix \texttt{clean, extremely detailed, best quality, sharp, high-resolution}. At inference time, we keep the original spatial size whenever possible, upscale by a factor of 4, align image dimensions to a multiple of 32, and perform a single LinearDiT update. Candidate checkpoints are ranked by the provided MUSIQ-based selection script under the same evaluation protocol used for final reporting.

For deployment compression, we first merge LoRA into the backbone and then run prompt-aware structured pruning with the same prompt distribution as evaluation. In the default pruning recipe, calibration uses 400 steps, batch size 1, BF16 calibration, and a target transformer budget of 0.35B parameters, while always preserving the first and last transformer blocks. The compressed model is then validated against the unpruned model under the same full-image or patch protocol, and each monitored metric is required to stay within a 3\% drop threshold while satisfying the target parameter budget.

\section{Additional experimental results}
\subsection{DRealSR Quantitative Results}
\label{app:drealsr_quant}

DRealSR~\citep{wei2020component} contains 93 LR–HR pairs captured by five different cameras, providing a complementary real-image evaluation to RealSR. Table~\ref{tab:main_drealsr} shows that SANA-SR achieves the best SSIM, MUSIQ, and DISTS on DRealSR, ranks second on both CLIPIQA and NIQE, and maintains competitive PSNR. These results further support that compact one-step LinearDiT restoration generalizes across distinct real-world capture conditions.

\subsection{Additional visualization results}
\label{app:more_visual}

\begin{figure}[htb]
	\centering
	\includegraphics[width=0.91\linewidth]{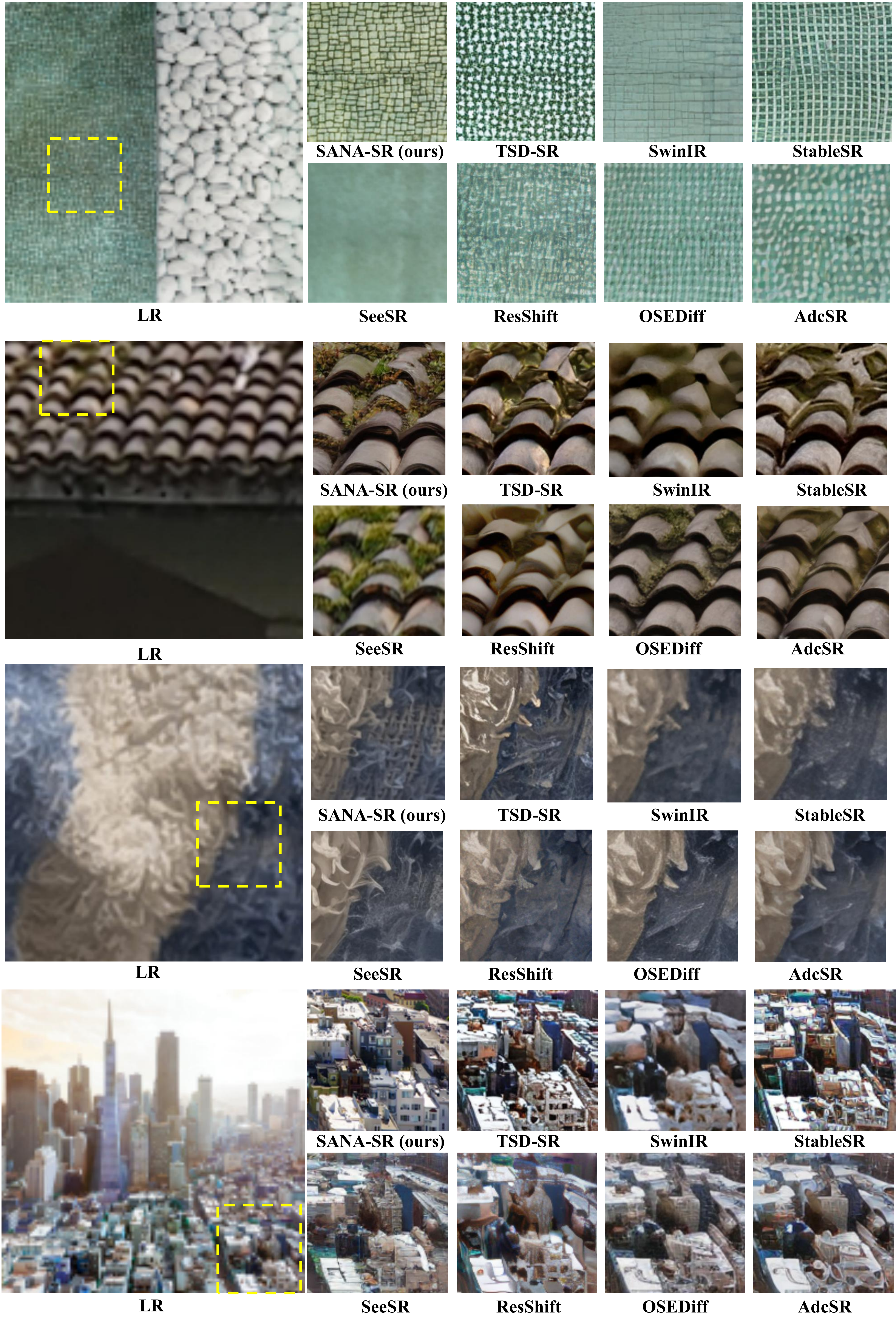}
    \caption{\textbf{Additional qualitative comparison} on examples from DIV2K-Val, RealSR, and DRealSR.}
    \label{fig:qualitative_appendix}
\end{figure}

Fig.~\ref{fig:qualitative_appendix} provides additional qualitative comparisons on challenging real-world images. In the urban scene, SANA-SR restores cleaner structural lines in the bridge arches, windows, and building boundaries, while maintaining more natural contrast in the water and facade regions. In the mechanical scene, our method better preserves the contours of the metal components and the fine boundaries around the bolts and rocker arms, avoiding the over-smoothed appearance observed in several baselines. These examples further support that SANA-SR can recover sharper local structures while maintaining visually coherent textures under complex real degradations.

\section{Further Analysis of Compact One-Step Restoration}
\label{app:further_analysis}

The results in the main paper establish the final restoration quality and efficiency of SANA-SR. This section examines why the operating point is effective. We first separate compact-representation stability from information recoverability, then study how the two prior-preserving objectives guide one-step adaptation, and finally analyze prompt conditioning, structured pruning, and execution cost. Together, these controls support the design principle underlying SANA-SR: an efficient one-step restorer benefits from reducing latent density and backbone depth while preserving task-specific guidance throughout adaptation and compression.

\subsection{Why the $32\times$ Latent Is Effective}
\label{app:compact_evidence}

We begin with the representation itself, independently of the final SR network. For a frozen encoder $E$ and a clean/degraded pair $(x,x_d)$, let $\Delta z=E(x_d)-E(x)$. We quantify the total encoded footprint of this residual on 92 independent DIV2K-Val source images, using disjoint 46/46 calibration and evaluation splits. Each residual channel is standardized using clean-channel statistics from the calibration split. A scalar probability table fitted on quantized calibration residuals estimates held-out cross-entropy; zlib-6 compression of the same arrays provides an independent redundancy-aware measurement. The SANA $32\times$/32-channel code contains 8,192 scalars, while the SDXL $8\times$/4-channel code contains 16,384. Confidence intervals use 10,000 paired source-level bootstrap resamples.

\begin{table}[htbp]
\centering
\small
\caption{Total degradation-residual footprint of the $32\times$ DC-AE relative to an $8\times$ VAE. Ratios below one indicate a smaller disturbance presented to the restoration network; brackets show paired 95\% confidence intervals.}
\label{tab:app_degradation_footprint}
\begin{tabular}{lcc}
\toprule
Degradation & Factorized rate ratio & zlib byte ratio \\
\midrule
Gaussian blur ($\sigma=1.6$) & 0.502 [0.494, 0.509] & 0.499 [0.493, 0.506] \\
Gaussian noise ($\sigma=15/255$) & 0.509 [0.505, 0.513] & 0.499 [0.495, 0.503] \\
JPEG compression (quality 50) & 0.513 [0.507, 0.519] & 0.503 [0.494, 0.512] \\
$4\times$ resize & 0.502 [0.496, 0.508] & 0.497 [0.492, 0.504] \\
Mixed degradation & 0.534 [0.527, 0.541] & 0.536 [0.531, 0.542] \\
\midrule
Macro average & \textbf{0.512 [0.508, 0.517]} & \textbf{0.508 [0.503, 0.513]} \\
\bottomrule
\end{tabular}
\end{table}

The compact code carries 48.8\% less factorized residual rate and 49.2\% fewer compressed residual bytes. This finding remains stable at quantization steps 0.05, 0.10, and 0.20, for which the factorized reductions are 49.0\%, 48.8\%, and 48.5\%, respectively. The measurement concerns total disturbance rather than per-coordinate invariance: the rate per scalar is comparable between codecs, while the compact code exposes fewer residual scalars to the restorer.

A smaller residual footprint and recoverable image detail are complementary properties. We therefore measure the frozen $32\times$ DC-AE reconstruction channel on 3,000 DIV2K patches from 92 sources. The source-balanced LPIPS is 0.0875 [0.0793, 0.0958]. For grayscale reference and reconstructed images, we additionally compare high-frequency energy after a $3\times3$ Laplacian filter $H$:
\begin{equation}
\epsilon_{\mathrm{HF}}
=\left|
\frac{\operatorname{RMS}(H(D(E(x))))}{\operatorname{RMS}(H(x))}-1
\right|.
\end{equation}
The resulting deviation is 3.43\% [3.04\%, 3.87\%]. Thus, the frozen codec substantially reduces the aggregate degradation footprint while retaining the measured perceptual content and high-frequency energy.

We next examine the end-task quality--resource trade-off. Native-$8\times$, Native-$32\times$, and Native-$64\times$ use their corresponding frozen codecs. A native $16\times$ codec is unavailable; the $16\times$ entry below is an invertible depth-to-space rearrangement of the $32\times$ latent, providing a token-layout control. The restoration variants share the 28-block LinearDiT, rank-64 LoRA, losses, data, and schedule.

\begin{table}[htbp]
\centering
\small
\caption{Quality and training resources across compact-latent settings. Quality is measured on RealSR. Resource measurements use equal-information token-grid controls that contain the same 8,192 latent scalars.}
\label{tab:app_compression_tradeoff}
\begin{tabular}{lrrrrrr}
\toprule
Setting & Tokens & PSNR $\uparrow$ & LPIPS $\downarrow$ & MUSIQ $\uparrow$ & FLOPs/step $\downarrow$ & Memory $\downarrow$ \\
\midrule
$8\times$ & 4,096 & 26.15 & 0.2897 & 71.23 & 32.86T & 41.62 GiB \\
$16\times$ grid & 1,024 & 26.07 & 0.2909 & 70.78 & 12.22T & 17.39 GiB \\
$32\times$ & 256 & 25.96 & 0.2917 & 70.73 & 7.06T & 11.15 GiB \\
$64\times$ & 64 & 22.13 & 0.3399 & 60.57 & 5.77T & 9.62 GiB \\
\bottomrule
\end{tabular}
\end{table}

Moving from the $16\times$ control to $32\times$ reduces counted training FLOPs by 42.20\% and allocated memory by 35.88\%, with changes of 0.11 dB PSNR and 0.05 MUSIQ. Moving from $32\times$ to the native $64\times$ system saves a further 18.26\% counted FLOPs and 13.72\% memory, while changing PSNR by 3.83 dB and MUSIQ by 10.16. At the submitted 16-block width, the $32\times$ grid also reduces Transformer MACs by 92.70\% relative to an $8\times$ grid. These measurements locate $32\times$ at a favorable quality--resource knee.

The same tendency appears under changing degradation strength and operation order. For every image in DIV2K-Val, RealSR, and DRealSR (3,193 images in total), we generate mild and strong observations under two operation orders, then compute the within-source relative standard deviation and range. Keeping the native $32\times$ information fixed and changing only its token-grid layout yields MUSIQ relative standard deviations of 0.1890, 0.1881, and 0.1793 for the $8\times$, $16\times$, and $32\times$ grids; the corresponding CLIPIQA values are 0.2514, 0.2490, and 0.2010. The compact layout therefore reduces output variation as well as token cost.

For completeness, ``linear attention'' refers to scaling in the spatial token count. Forming $\phi(K)^\top V$ and contracting it with $\phi(Q)$ each cost $\mathcal O(Nd^2)$ when the value width is proportional to hidden width $d$, rather than the $\mathcal O(N^2d)$ cost of quadratic self-attention. With fixed width, text length, and depth, SANA-SR therefore scales linearly with $N$; the exact expression retains the $d^2$ factor.

\subsection{Target-Anchored Prior Adaptation}
\label{app:prior_adaptation}

The alignment and consistency objectives preserve different aspects of the pretrained prior. In alignment, the model is fixed and the input changes from the restored latent to the paired HQ latent:
\begin{equation}
f_0(\tilde z_{\hat z},t,c,m)
\quad\text{and}\quad
f_0(\tilde z_H,t,c,m).
\end{equation}
In consistency, the input is fixed and the model changes from the adapted to the frozen backbone:
\begin{equation}
f_\theta(\tilde z_{\hat z},t,c,m)
\quad\text{and}\quad
f_0(\tilde z_{\hat z},t,c,m).
\end{equation}
Consequently, alignment is target-relative, whereas consistency is adapter-drift-relative. Both alignment branches share the same $(t,\epsilon,c,m)$, so their perturbed-input difference simplifies to $\alpha_t(\hat z-z_H)$; matched noise removes the nuisance term that would arise from independent perturbations.

The paired HQ latent supplies the destination in this fixed response space. To test its specificity, we keep the restored query, $f_0$, prompt, timestep, noise, and preprocessing unchanged, replacing only the correct HQ reference with 20 randomly mismatched references from the same dataset. We evaluate all 100 RealSR and 93 DRealSR pairs at three seeded timesteps, obtaining 579 probes and 11,580 comparisons.

\begin{table}[htbp]
\centering
\small
\caption{Content specificity of frozen-prior alignment. $d^+$ and $d^-$ denote the distances to paired and mismatched HQ references. Confidence intervals use 10,000 image-level bootstrap resamples.}
\label{tab:app_alignment_specificity}
\begin{tabular}{lrrrr}
\toprule
Dataset & $d^+$ $\downarrow$ & $d^-$ $\downarrow$ & Relative reduction & $\Pr(d^+<d^-)$ \\
\midrule
RealSR & 0.0262 & 0.1221 & 78.49\% [73.88, 82.15] & 94.87\% [92.38, 96.87] \\
DRealSR & 0.0342 & 0.2390 & 84.30\% [82.37, 86.10] & 97.29\% [96.22, 98.23] \\
Overall & 0.0301 & 0.1784 & 81.29\% [78.70, 83.49] & 96.04\% [94.63, 97.19] \\
\bottomrule
\end{tabular}
\end{table}

The paired reference wins in 11,121 of 11,580 comparisons, and the image-averaged ordering holds for 99.48\% of queries. The frozen backbone therefore provides a response geometry while the paired HQ image determines the content-specific target. Direct pixel and LPIPS supervision anchors the output in image space at the same time.

We further vary matched noise and the alignment anchor on RealSR. Independent-noise HQ alignment gives PSNR/LPIPS/MUSIQ of 25.43/0.3109/68.91, while matched-noise LQ alignment gives 24.68/0.3387/65.84. Matched-noise HQ alignment obtains 25.96/0.2917/70.73. Sharing the perturbation improves response comparison, and the paired HQ anchor improves both fidelity and perceptual quality.

The complete factorial ablation in Table~\ref{tab:app_loss_factorial} changes only the two regularizer weights while retaining pixel and LPIPS supervision. Each objective improves all three task metrics both independently and in the presence of the other.

\begin{table}[htbp]
\centering
\small
\caption{Factorial RealSR ablation of frozen-prior alignment and adapter consistency.}
\label{tab:app_loss_factorial}
\begin{tabular}{lccrrr}
\toprule
Objective & $\lambda_a$ & $\lambda_c$ & PSNR $\uparrow$ & LPIPS $\downarrow$ & MUSIQ $\uparrow$ \\
\midrule
Task supervision & 0 & 0 & 21.95 & 0.4257 & 57.38 \\
With consistency & 0 & 1 & 24.14 & 0.3540 & 66.74 \\
With alignment & 1 & 0 & 24.77 & 0.3410 & 67.90 \\
Full objective & 1 & 1 & \textbf{25.96} & \textbf{0.2917} & \textbf{70.73} \\
\bottomrule
\end{tabular}
\end{table}

Adding alignment to consistency improves PSNR by 1.82 dB, LPIPS by 0.0623, and MUSIQ by 3.99. Adding consistency to alignment improves the same metrics by 1.19 dB, 0.0493, and 2.83. These favorable conditional effects show that the losses provide complementary constraints at the adopted weights.

\subsection{Prompt Conditioning and Structured Compression}
\label{app:prompt_pruning_analysis}

The visual latent and text condition serve complementary roles. The latent $z_L=E(x_L)$ supplies spatial queries, while $(c,m)=\mathcal T(\Pi(x_P)\oplus p_{\mathrm{tpl}})$ supplies semantic and quality keys/values and marks valid token positions. Under the HQ-prompt protocol used for the main quantitative results, $x_P=x_H$; the released input-only protocol instead uses $x_P=x_L$. Across all 3,193 benchmark images, replacing HQ-DAPE conditioning with an empty encoded prompt changes MUSIQ by 4.31, 3.07, and 4.13 on DIV2K, RealSR, and DRealSR, and changes CLIPIQA by 0.117, 0.085, and 0.110, respectively. The condition therefore provides consistent semantic and clean-quality guidance.

The tokenizer mask has a narrower function: it excludes padding positions from text cross-attention. Replacing it with an all-one mask while keeping the prompt fixed changes image-weighted CLIPIQA from 0.6919 to 0.6917 over all 3,193 images. A representative prompt with 40 valid and 260 padded positions shows a larger local response: applying the tokenizer mask changes PSNR from 19.1074 to 19.5735, LPIPS from 0.4445 to 0.4239, MUSIQ from 72.81 to 74.11, and CLIPIQA from 0.7022 to 0.7279. Thus, semantic content comes from the prompt, while $m$ preserves valid-token attention.

After adaptation, prompt-aware pruning selects 16 of the original 28 blocks. To isolate the selector itself, we compare four whole-block criteria from the same 591.748M-parameter parent, with the same merged LoRA, calibration data, prompt protocol, boundary constraint, and exact 343.994M-parameter budget.

\begin{table}[htbp]
\centering
\small
\caption{Matched-budget structured block selection. DISTS is evaluated on the same 100 RealSR and 93 DRealSR pairs; lower is better.}
\label{tab:app_pruning_selectors}
\begin{tabular}{lrrr}
\toprule
Selector & RealSR & DRealSR & Aggregate \\
\midrule
Prompt-aware OBS & \textbf{0.2734} & \textbf{0.2765} & \textbf{0.2749} \\
SLEB-style greedy elimination & 0.2785 & 0.2910 & 0.2846 \\
Block-influence ranking & 0.2798 & 0.2926 & 0.2860 \\
First-order Taylor & 0.2856 & 0.2933 & 0.2893 \\
\bottomrule
\end{tabular}
\end{table}

Prompt-aware selection improves aggregate DISTS by 3.40\%, 3.88\%, and 4.98\% relative to the three generic criteria. Its retained zero-based indices are $\{0,11\text{--}24,27\}$, whereas first-order Taylor retains $\{0,13\text{--}27\}$. Retaining blocks 11 and 12 while removing blocks 25 and 26 indicates restoration-conditioned interior selection beyond a fixed late-block rule. The first and last blocks are kept in every comparison. Removing both from the submitted 16-block model without retraining changes MUSIQ by $-29.91$ on RealSR and $-23.76$ on DRealSR, supporting this boundary constraint.

\subsection{Efficiency Accounting and Deployment Scope}
\label{app:efficiency_scope}

The 0.019 s, 407.95 GMAC, and 344M values in the main paper characterize the pruned DiT backbone. The original profile uses a representative $992\times1408$ working image. In the same ten-image run, the pruned restoration core (DC-AE encode, one DiT update, and DC-AE decode) averages 0.322 s, and the prompt-inclusive path averages 0.395 s. The corresponding dense-model times are 0.358 s and 0.430 s. These measurements separate the substantial DiT reduction from the remaining codec and prompt-processing cost.

We also benchmark official runnable pipelines on the same deterministic $256\times256$ input and $1024\times1024$ output with batch size one on an RTX PRO 6000. Each measurement uses three warm-ups, ten synchronized repetitions, and official precision. Model loading and file I/O are excluded.

\begin{table}[htbp]
\centering
\small
\caption{Controlled restoration-pipeline efficiency. Methods retain their official sampling and conditioning settings. MACs use supported profiler operators and are reported as FLOPs/2.}
\label{tab:app_pipeline_efficiency}
\begin{tabular}{lrrr}
\toprule
Method & Steps & Latency (s) $\downarrow$ & MACs (T) $\downarrow$ \\
\midrule
SANA-SR restoration core & 1 & $0.1444\pm0.0005$ & 7.087 \\
TSD-SR & 1 & $0.5679\pm0.0039$ & 16.439 \\
InvSR & 1 & $0.2991\pm0.0013$ & 9.843 \\
LinearSR & 20 & $1.0324\pm0.0565$ & 63.038 \\
DreamClear & 50 & $32.8176\pm0.5081$ & 1,370.377 \\
\bottomrule
\end{tabular}
\end{table}

The closest one-step comparison is SANA-SR against TSD-SR: the SANA-SR core is 3.93$\times$ faster and uses 56.89\% fewer counted MACs. InvSR, LinearSR, and DreamClear represent inversion, multi-expert, and caption-conditioned pipelines, respectively, so their sampling depth and conditioning remain explicit rather than being collapsed into one homogeneous ranking.

Finally, we export the complete 16-block, 343.994M-parameter pruned DiT with static QOperator INT8 Conv/MatMul/Gemm operators and FP16 inputs/outputs. On a Huawei Mate 70 Pro CPU using ONNX Runtime 1.23.2 and eight threads, one DiT forward plus latent update takes $578.38\pm18.63$ ms after three warm-ups over ten runs. Peak sampled PSS is 587.33 MiB and the model file is 332.75 MiB. The execution profile records 22,178 CPU-provider node events. This experiment demonstrates sub-second execution of the quantized restoration backbone; DC-AE encoding/decoding, prompt extraction/encoding, and image I/O remain outside the phone timing boundary.

\subsection{Behavior on Challenging Structures}
\label{app:failure_behavior}

The dense and pruned models are also evaluated on the same RealSR and DRealSR pairs to identify the image structures that remain most demanding. The dense one-step prior obtains relatively low perceptual scores on repeated building/window patterns (RealSR/Canon\_041), fine blossoms (RealSR/Canon\_015), and dense flowers or vegetation (DRealSR/DSC\_1575). These cases share many small, repeated elements whose exact local arrangement is weakly constrained by the LQ observation. This observation is consistent with the qualitative comparisons already provided in Figs.~\ref{fig:qualitative_results} and~\ref{fig:qualitative_appendix}: compact one-step restoration preserves salient structure well, while highly repetitive details remain the most sensitive content class.

Matched dense/pruned evaluation further reveals where block removal has the largest local effect. On RealSR/Canon\_029, Nikon\_005, and Nikon\_007, pruning changes MUSIQ by $-15.77$, $-10.15$, and $-15.81$, and CLIPIQA by $-0.244$, $-0.070$, and $-0.086$, respectively. These examples motivate a natural extension of prompt-aware selection: content-balanced calibration can place additional emphasis on repeated geometry and fine vegetation while retaining the same global parameter budget.

\section{Limitations and Future Work}
\label{app:limitations}

While SANA-SR achieves a strong quality--efficiency trade-off, several limitations remain. (1) Our experiments use the Real-ESRGAN-style degradation pipeline; broader real-world degradations remains to be explored. (2) The reported $0.019$ s inference is measured on a single NVIDIA RTX 4090 and may differ on mobile NPUs without additional quantization. (3) We focus on $4\times$ super-resolution; ultra-high upscaling factors (e.g., $8\times$, $16\times$) likely require further architectural adjustments.


\end{document}